\pdfoutput=1

\documentclass[11pt]{article}

\usepackage{ACL2023}

\usepackage{times}
\usepackage{latexsym}

\usepackage[T1]{fontenc}

\usepackage[utf8]{inputenc}

\usepackage{microtype}

\usepackage{inconsolata}
\usepackage{multirow}
\usepackage{multicol}
\usepackage{graphicx}
\usepackage{subcaption}
\usepackage{todonotes}
\usepackage{booktabs}
\usepackage{amsmath}
\usepackage{tipa}

\title{
WebVoyager \includegraphics[width=0.04\textwidth]{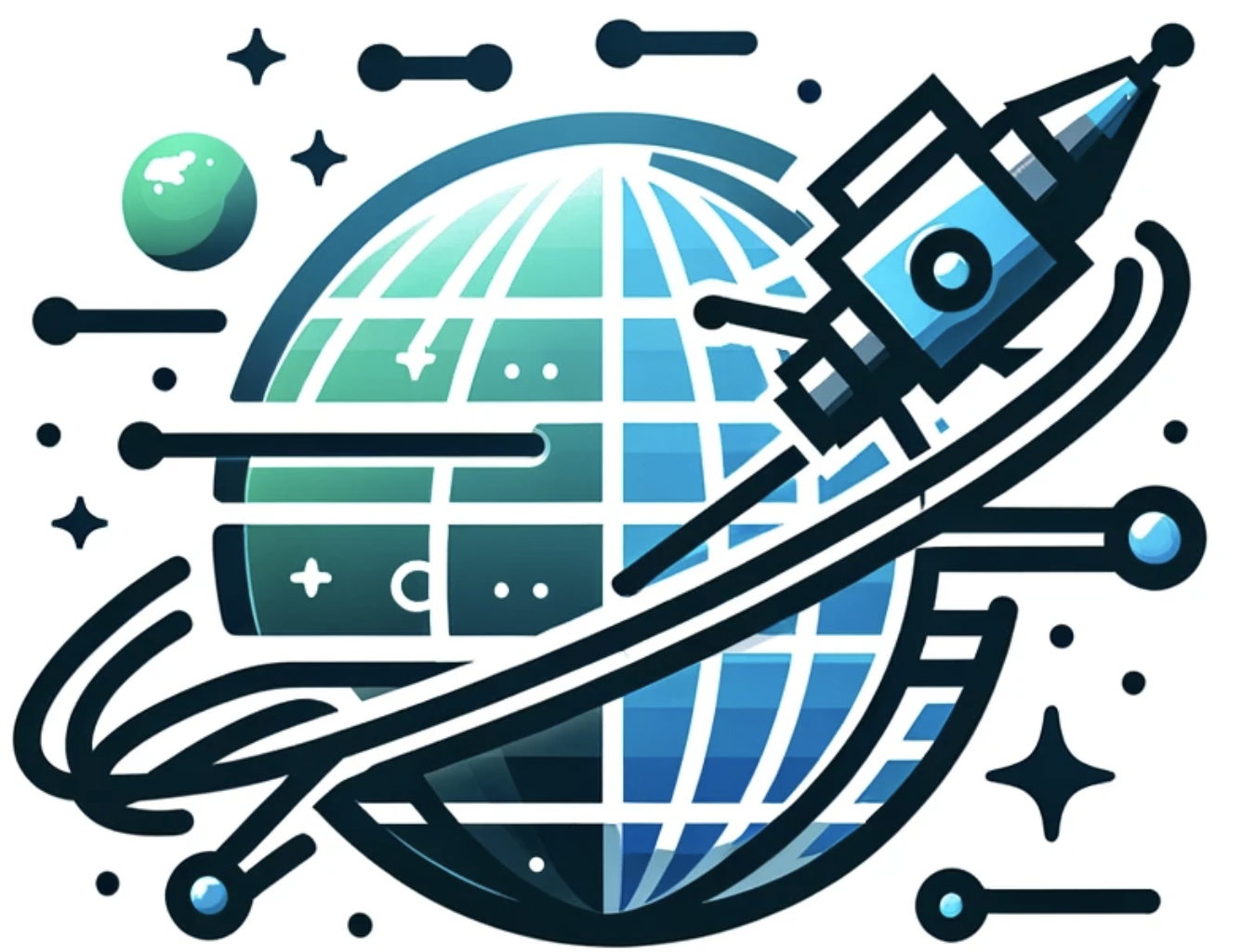}:
Building an End-to-End Web Agent with 
\\ Large Multimodal Models 
}

\author{Hongliang He$^{1,3}$\thanks{\ \ Work done during the internship at Tencent AI Lab.} , 
        Wenlin Yao$^{2}$, 
        Kaixin Ma$^{2}$, 
        Wenhao Yu$^{2}$, 
        Yong Dai$^{2}$, \\ 
        \textbf{Hongming Zhang}$^{2}$,
        \textbf{Zhenzhong Lan}$^{3}$,
        \textbf{Dong Yu}$^{2}$ \\
        $^{1}$Zhejiang University,
        $^{2}$Tencent AI Lab, 
        $^{3}$Westlake University \\
        \texttt{hehongliang@westlake.edu.cn}, \texttt{wenlinyao@global.tencent.com}
    }

\begin{document}
\maketitle
\begin{abstract}

The rapid advancement of large language models (LLMs) has led to a new era marked by the development of autonomous applications in real-world scenarios, which drives innovation in creating advanced web agents.
Existing web agents typically only handle one input modality and are evaluated only in simplified web simulators or static web snapshots, greatly limiting their applicability in real-world scenarios. 
To bridge this gap, we introduce WebVoyager, an innovative Large Multimodal Model (LMM) powered web agent that can complete user instructions end-to-end by interacting with real-world websites. 
Moreover, we establish a new benchmark by compiling real-world tasks from 15 popular websites and introduce an automatic evaluation protocol leveraging multimodal understanding abilities of GPT-4V to evaluate open-ended web agents.
We show that WebVoyager achieves a 59.1\% task success rate on our benchmark, significantly surpassing the performance of both GPT-4 (All Tools) and the WebVoyager (text-only) setups, underscoring the exceptional capability of WebVoyager.
The proposed automatic evaluation metric achieves 85.3\% agreement with human judgment, 
indicating its effectiveness in providing reliable and accurate assessments of web agents.
\footnote{Our code and data will be released at \url{https://github.com/MinorJerry/WebVoyager}}

\end{abstract}

\section{Introduction}

The recent advancement of large language models (LLMs), such as ChatGPT and GPT-4 \citep{openai2023gpt4}, have sparked significant interest in developing LLM-based autonomous agents~\citep{Significant_Gravitas_AutoGPT} for complex task execution \citep{qin2023toolllm, schick2023toolformer}. 
Recent studies have explored the construction of text-based web browsing environments and how to instruct large language model agents to perform web navigation \citep{nakano2021webgpt, gur2023real, zhou2023webarena, lu2023chameleon}. The primary challenge in these works lies in managing complex and verbose HTML texts, and solutions include simplifying and structuring HTML~\citep{nakano2021webgpt, zhou2023webarena,gur2023real,deng2023mind2web}.

However, existing approaches overlook a critical functionality of browsing: rendering HTML into visual webpages. 
Particularly, vision capability is crucial for utilizing tools such as web browsers, as rendered web pages are inherently designed with user experience (UX), emphasizing intuitive information and structured presentation. This design principle of rendering makes visual analysis more effective than mere HTML representation.
At present, large multimodal models (LMMs), particularly GPT-4V(ision) \citep{openai2023gpt4} and Gemini \citep{team2023gemini}, demonstrate a remarkable ability to integrate intricate visual cues with textual information. 
Existing studies such as Pix2Struct \citep{lee2023pix2struct} and WebArena \citep{zhou2023webarena}, have initiated explorations into using screenshots as inputs for decision-making in web navigation, yet these are preliminary and do not represent a deep exploration.
Therefore, building multimodal web agents to leverage the environment rendered by browsers through screenshots, thus mimicking human web browsing behavior, is now a viable approach to enhance web navigation abilities.

\begin{figure*}[t]
\centering
\includegraphics[width=0.8\linewidth]{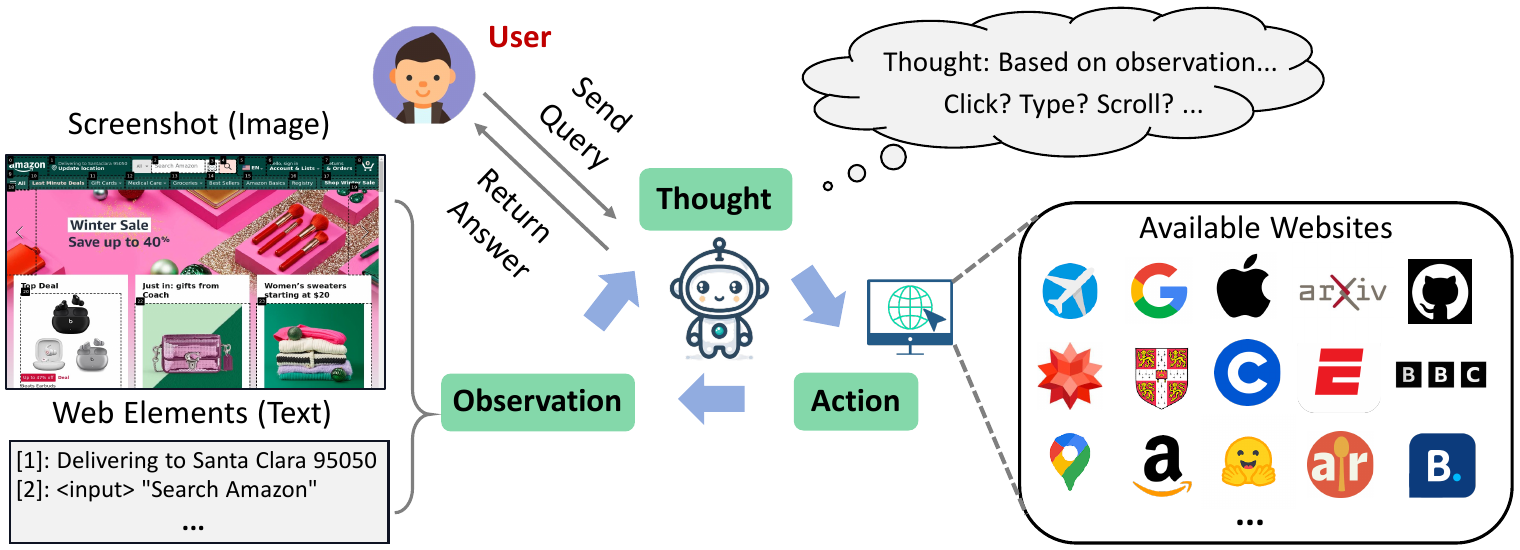}
\caption{The overall workflow of WebVoyager. 
WebVoyager takes web tasks assigned by a human and automatically browses the web online.
At each step, WebVoyager selects actions based on screenshots and text (the `type' of the web element and its contents). Once the task is completed, the answers will be returned to the user. For example, for a user query: "Find the cost of a 2-year protection for PS4 on Amazon.", the agent interacts with Amazon online, locates the PS4, identifies the 2-year protection price, and returns "\$30.99" to the user.}
\vspace{-0.1in}
\label{fig:overall}
\end{figure*}

We introduce WebVoyager (Figure \ref{fig:overall}), a multimodal web agent designed to autonomously accomplish web tasks online from start to finish, managing the entire process end-to-end without any intermediate human intervention.
WebVoyager processes the user query by making observations from screenshots and textual content in interactive web elements, formulates a thought on what action to take (such as clicking, typing, or scrolling, etc.), and then executes that action on the websites.
Inspired by Set-of-Mark Prompting \citep{yang2023set}, we mark interactive web elements on screenshots (see Figure \ref{fig:web_mark}) to facilitate decision-making for WebVoyager. 

Another challenge is the evaluation of an end-to-end web agent. 
Existing benchmarks, such as Mind2Web \citep{deng2023mind2web}, primarily focus on stepwise and offline evaluation, where agents follow a predefined ``golden'' trajectory for action selection. 
This approach, however, may not fully account for the variety of viable strategies to accomplish a task, as it only reflects one possible plan. This limitation could lead to a biased evaluation and difficulties in fairly comparing different methods.
To accurately evaluate the capabilities of web agents in end-to-end task completion, we propose an automated evaluation protocol using GPT-4V. 
Specifically, we save screenshots throughout the online navigation process and then use GPT-4V to evaluate these trajectories together with the final results automatically.
Human evaluations are also conducted to verify the results and 
the analysis shows that our evaluation protocol achieves 85.3\% agreement with human judges, indicating GPT-4V can serve as a reliable evaluator for online agents.


We conduct evaluations on a newly collected dataset, which is semi-automatically generated using a self-instruct \citep{wang2022self} method, comprising 643 web tasks from 15 commonly accessed websites. We also evaluate WebVoyager on 90 web-related tasks of level 1 and level 2 from the GAIA \citep{mialon2023gaia}, and 50 interactive open-web tasks from SeeAct \cite{zheng2024gpt}.  
We compare our WebVoyager with 1) GPT-4 (All Tools)\footnote{GPT-4 (All Tools) is an integrated tool-based agent 
released by OpenAI in Oct. 2023. See https://chat.openai.com/}, and 2) WebVoyager in a text-only setting which employs the textual accessibility tree proposed in WebArena  \citep{zhou2023webarena} to describe web pages. 
The results show that WebVoyager achieves a Task Success Rate of 59.1\% on our new benchmark, significantly outperforming GPT-4 (All Tools) with a rate of 30.8\% and the text-only setting with a rate of 40.1\%, demonstrating the effectiveness of our method. 
Our research demonstrates the effectiveness of the WebVoyager method for web tasks, offering insights into the development of more intelligent and efficient web automation solutions.


    
     

\section{Related Work}
Autonomous web navigation \citep{shi2017world, yang2023dawn} requires an agent to follow instructions, construct plans, comprehend complex web structures, and decompose tasks into step-by-step decisions \citep{weng2023prompt}. To study web agents in a controlled environment, previous works constructed web simulators that contain simplified websites \citep{shi2017world,yao2022webshop}. More recently, there has been a surge of interest in building more challenging and realistic benchmarks such as Mind2Web \cite{deng2023mind2web} and WebArena \cite{zhou2023webarena}. 

Along with these new benchmarks, numerous efforts have been made to build autonomous web agents. WebGPT \citep{nakano2021webgpt} constructs a text-based web browsing environment and fine-tunes GPT-3 as a web agent. WebAgent \citep{gur2023real} pretrains a T5 model to extract HTML snippets and leverages Flan-U-PaLM \citep{chowdhery2023palm} to generate Python code to interact with the environment. Besides fine-tuning, another line of work tries to build web agents by prompting LLMs \cite{yao2022react, shinn2023reflexion, ma2023laser}. 
Multimodal web agents that integrate visual signals have also been explored, WebGUM \citep{furuta2023multimodal} combines T5 \citep{raffel2020exploring} with a Vision Transformer (ViT) to navigate using both screenshots and HTML text. PIX2ACT \citep{shaw2023pixels} instead solely relies on web screenshots as inputs to predict agent actions. Unlike previous works that only consider a single modality or simplified web environments, we build a multimodal agent that can complete tasks on real-world websites in this work.
Concurrently with our work, SeeAct \citep{zheng2024gpt} also leverages Large Multimodal Models (LMMs) for integrated visual understanding and actions on websites. However, the best SeeAct agent still relies on a finetuned cross-encoder model to select candidate elements for interaction. In contrast, WebVoyager does not require any additional modules.  

\section{WebVoyager} 
We aim to build an agent that can browse the open web autonomously without human intervention to complete user instructions. Given an instruction, our WebVoyager first instantiates a web browser and then performs actions with visual (i.e., screenshots) and textual (i.e., HTML elements) signals from the web. 
The agent produces an action based on the inputs at every step, which is then executed in the browser environment. 
The process continues until the agent decides to stop. The details of WebVoyager, including environment, interaction cycle, observation space, and action space, are as follows.

\subsection{Browsing Environment}
We develop an automated web-browsing environment using Selenium\footnote{https://www.selenium.dev/}. 
Unlike WebArena~\cite{zhou2023webarena}, we do not host any websites locally and allow the agent to explore the open web instead, which poses unique challenges such as floating ads, pop-up windows, constant updates, etc.\footnote{Regarding CAPTCHAs (Completely Automated Public Turing test to tell Computers and Humans Apart) challenges, we believe it is important to respect the rules of these websites and prompt the agent to retrieve information from alternative sources.} Still, we opt for online interaction with real websites as we believe that this setting truly reflects the real-world use cases (e.g., the agent needs access to real-time information from the web), and a successful web agent should be able to adapt to these challenges and consistently solve the problem robustly. 


\subsection{Interaction Formulation}\label{subsec:planning_method}

Formally, we denote the Environment as \(\mathcal{E}\), the large Multimodal Model as \(\mathcal{M}\), the Observation Space as \(\mathcal{O}\), and the Action Space as \(\mathcal{A}\). At time step $t$, the model receives the context $c_t$ as inputs, 
which consist of historical actions $a_i$ and observations $o_i$, defined as: $c_t = (o_1, a_1, ..., o_{t-1}, a_{t-1}, o_t, I)$ 
The the model produces the action $a_t$ at time 
$t$,
$a_t = \mathcal{M}(c_t)$, which is then executed in the environment. After execution, the environment sends back the observation at time $t+1$, $o_{t+1} = \mathcal{E}(o_t, a_t)$. Then the context will be updated and this interaction process continues until the model generates a terminating action or the maximum step is reached. 


Inspired by the paradigm of ReAct Prompting 
\citep{yao2022react}, we also prompt our agent to generate a thought process first before generating the action code. Hence $a_t$ can be further composed into $(s_t, \hat{a}_t)$ where $s_t$ and $\hat{a}_t$ represent the natural language thought and action code respectively. 
Figure \ref{fig:prompt} in Appendix \ref{sec:appendixA} presents the System Prompt we designed for the action prediction step. Also, it's worth noting that excessive observations of web pages from longer episodes may confuse the agent. Therefore, we perform context clipping to remove outdated web page information and only keep the three most recent observations in the inputs, and we keep the entire history of thoughts and actions to better guide the agent. 

\begin{figure}[t!]
\centering
\includegraphics[width=0.92\linewidth]{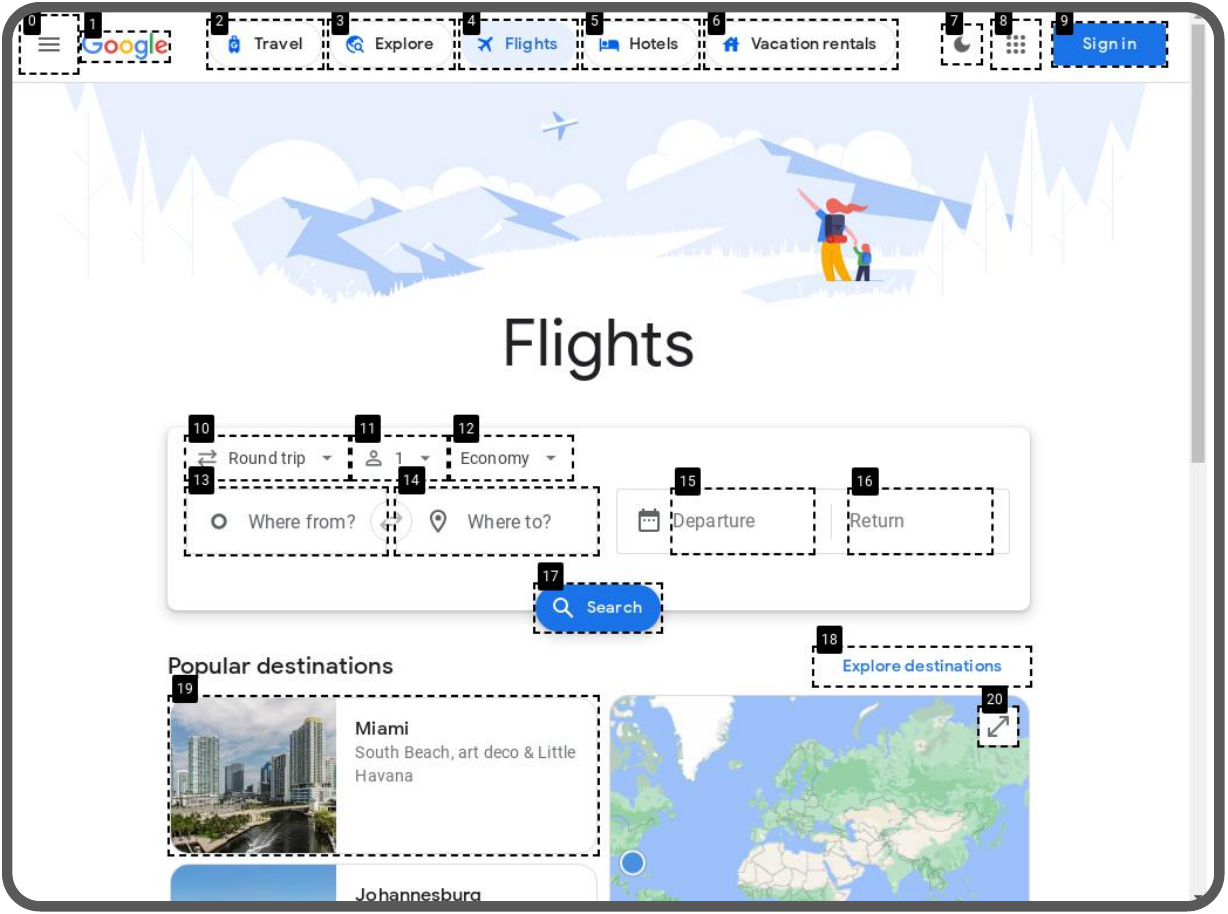}
\caption{Examples of webpage screenshots provided to the agent. We add borders to most of the interactive elements on the web pages and label them with numerical tags in the top left corner.}
\label{fig:web_mark}
\vspace{-0.1in}
\end{figure}

\subsection{Observation Space}
Similar to how humans browse the web, our agent also takes the visual information from the web (screenshots) as the primary source of input. Using screenshots allows us to avoid the burden of processing HTML DOM tree or accessibility tree to portray the overall structure of webpages, which can lead to overly verbose texts and impact the decision-making process of the agent. Inspired by Set-of-Mark Prompting \citep{yang2023set}, we overlay bounding boxes of the interactive elements on the websites to better guide the agent's action prediction. Unlike \citet{yang2023set}, we do not need any object detection module \citep{zou2023object}. Instead, we utilize 
GPT-4V-ACT\footnote{https://github.com/ddupont808/GPT-4V-Act}, a Javascript tool to extracts the interactive elements based on web element types and then overlays bounding boxes with numerical labels on the respective regions of the elements. GPT-4V-Act is efficient since it is rule-based without incorporating any object detection model. 

As illustrated in Figure \ref{fig:web_mark}, the nature of webpages allows us to locate and outline each interactive element using this tool precisely. The numerical labels assigned to each element are also essential for the model to identify the elements requiring interaction, thereby facilitating accurate action determination. We empirically choose black color for the borders and the background of the labels to enhance clarity. We observe that using a single black color yields higher success rates than using multiple colors.
We also provide the agent with auxiliary text as inputs, including the textual content embedded within the interactive element, the type of the element, and possibly some comment text in the aria-label attribute. To simplify the observation, we have disabled multiple tabs, i.e., all interactions occur within the current tab instead of opening new ones. 

At every step, the agent receives the current screenshot, auxiliary text, and history as inputs, as discussed in (\S\ref{subsec:planning_method}). In case the agent's action raised an exception during execution, we additionally incorporated the error messages in the prompt and asked the model to regenerate the response. Note that each error correction attempt also consumes one step from the total exploration budget. 

\subsection{Action Space}
We define the action space of our agent similar to how humans browse the web. 
To this end, we implement the most commonly used mouse and keyboard actions, sufficient for the agent to browse various web pages and locate the content required for the task. With the help of numerical labels in screenshots, we enable the agent to respond with a concise Action Format. This approach precisely locates the elements requiring interaction and executes the corresponding actions. The usage of actions is as follows (more details in Appendix \ref{sec:action_space}): 
1) \textbf{Click.} This action involves clicking on an element within a webpage, typically a link or a button. 
2) \textbf{Input.} This composite action involves selecting a text box, deleting any existing content within it, and then inputting new content.
3) \textbf{Scroll}. Scrolling is a common operation for browsing webpages, usually involving the vertical movement of the entire page. 
4) \textbf{Wait}. Action execution requires time, and this action is often used to wait for web pages to load. 
5) \textbf{Back}. This action is used to return to the previous page. 
6) \textbf{Jump to Search Engine}. There are often situations where agents get stuck at a certain website without finding an answer. This action enables the agent to jump to a search engine and start anew. 
7) \textbf{Answer}. Once all questions in the task are resolved, this action concludes the iteration and provides an answer in line with the task requirements.

\section{Benchmark for WebVoyager}\label{sec:benchmark}

\subsection{Website Selection}
We select 15 representative websites that cover different aspects of our daily life to ensure diversity in our evaluation, including Allrecipes, Amazon, Apple, ArXiv, BBC News, Booking, Cambridge Dictionary, Coursera, ESPN, GitHub, Google Flights, Google Map, Google Search, Huggingface, and Wolfram Alpha. Due to technical limitations, we regretfully omit websites requiring login or CAPTCHA to access their content. Additionally, Google Search is a universal website that can serve as a starting point for any website, making our framework applicable to various scenarios.

\begin{figure}[t!]
\centering
\includegraphics[width=0.85\linewidth]{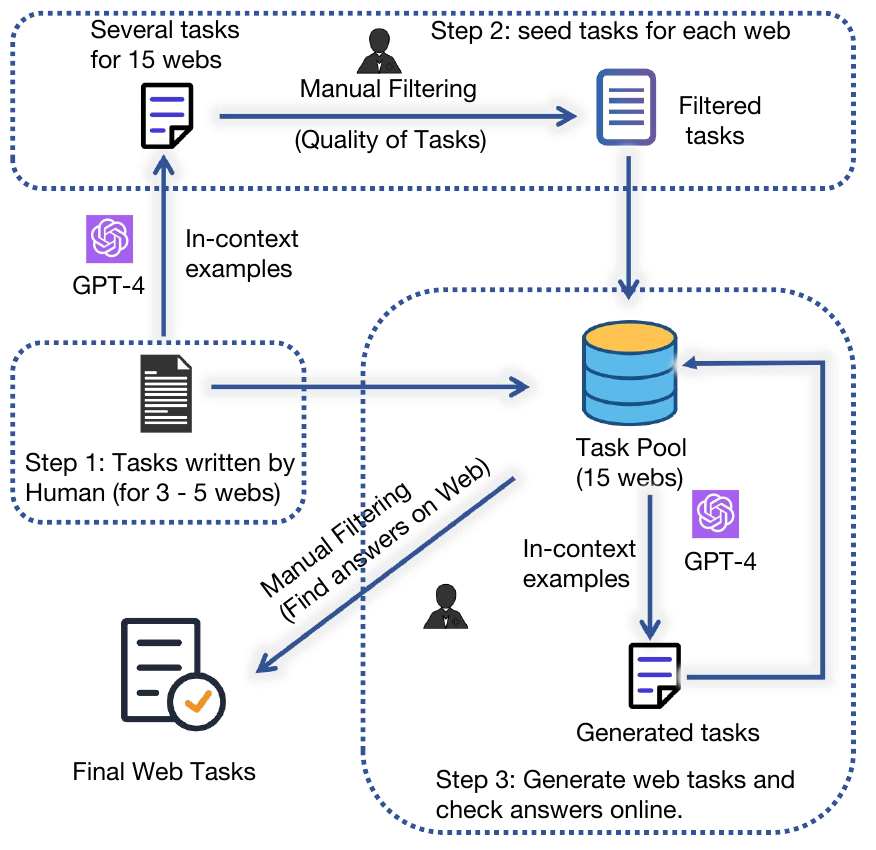}
\caption{Data creation process using self-instruct.}
\label{fig:data_creation}
\vspace{-0.1in}
\end{figure}

\subsection{Data Construction}
We employ a combination of self-instruct \citep{wang2022self} and human verification to construct our evaluation set. Figure \ref{fig:data_creation} illustrates our data creation process. Initially, we manually sample and rewrite some tasks from Mind2Web \citep{yin2023lumos, deng2023mind2web} for websites including Google Flights, Google Map, Google Search, Booking, and Wolfram Alpha. 
This process yields initial seed tasks in the Task Pool for subsequent generation. 
In step two, we sample tasks from Task Pool as in-context examples \citep{dong2022survey} and prompt GPT-4 Turbo to generate approximately 100 new tasks (20 iterations). Then we manually verify each generated task and rewrite them if necessary to ensure its high quality and the answers can be found on the corresponding website, then we add them to the Task Pool as additional seed tasks. This step allows us to create human-validated seed tasks for each website. 
Finally, in step three, we sample more diverse in-context examples in the Task Pool and directly add the generated tasks to the Task Pool in each iteration. We manually verify that the generated tasks have low repetition, and the answers to the generated tasks can be found on the websites.
In total, we collected 40+ tasks per website, resulting in a total of 643 tasks.

To further confirm that the generated tasks in the dataset have low repetition, We use the all-mpnet-base-v2\footnote{https://huggingface.co/sentence-transformers/all-mpnet-base-v2} model to calculate pairwise similarity for 643 ques. Out of a total of 206,403 pairs, only 49 pairs have a similarity greater than 0.8, and 140 pairs have a similarity between 0.7 and 0.8. All of these have been manually checked and found to be acceptable. 99.68\% of pairs have a similarity of less than 0.6. This demonstrates the diversity of our tasks and the robustness of our approach.

\subsection{Annotation Process}
After collecting the full task pool, we annotate answers for each task. Since some questions are open-ended and the web information may change, these questions may not have a fixed golden response. Thus, we label each data entry with an answer, categorized as ``Possible'' or ``Golden.'' For answers labeled as ``Golden,'' we provide a comprehensive listing of possible responses and consider them stable in the short term. The ``Possible'' category covers the following scenarios: 1) Answers for open-ended tasks where it's hard to find an exact match answer, such as summarization. 2) multiple answers satisfy the task, making it impractical to list all of them. Therefore, we provide a partial listing. 3) Tasks related to real-time information, where the answer might change, e.g., flight ticket prices. Hence, the ``Possible'' answers were also correct during our experiments. 
In total, 22.3\% of questions are annotated with golden responses, and the rest only have possible answers. 


\begin{figure*}[t!]
\centering
\includegraphics[width=0.85\linewidth]{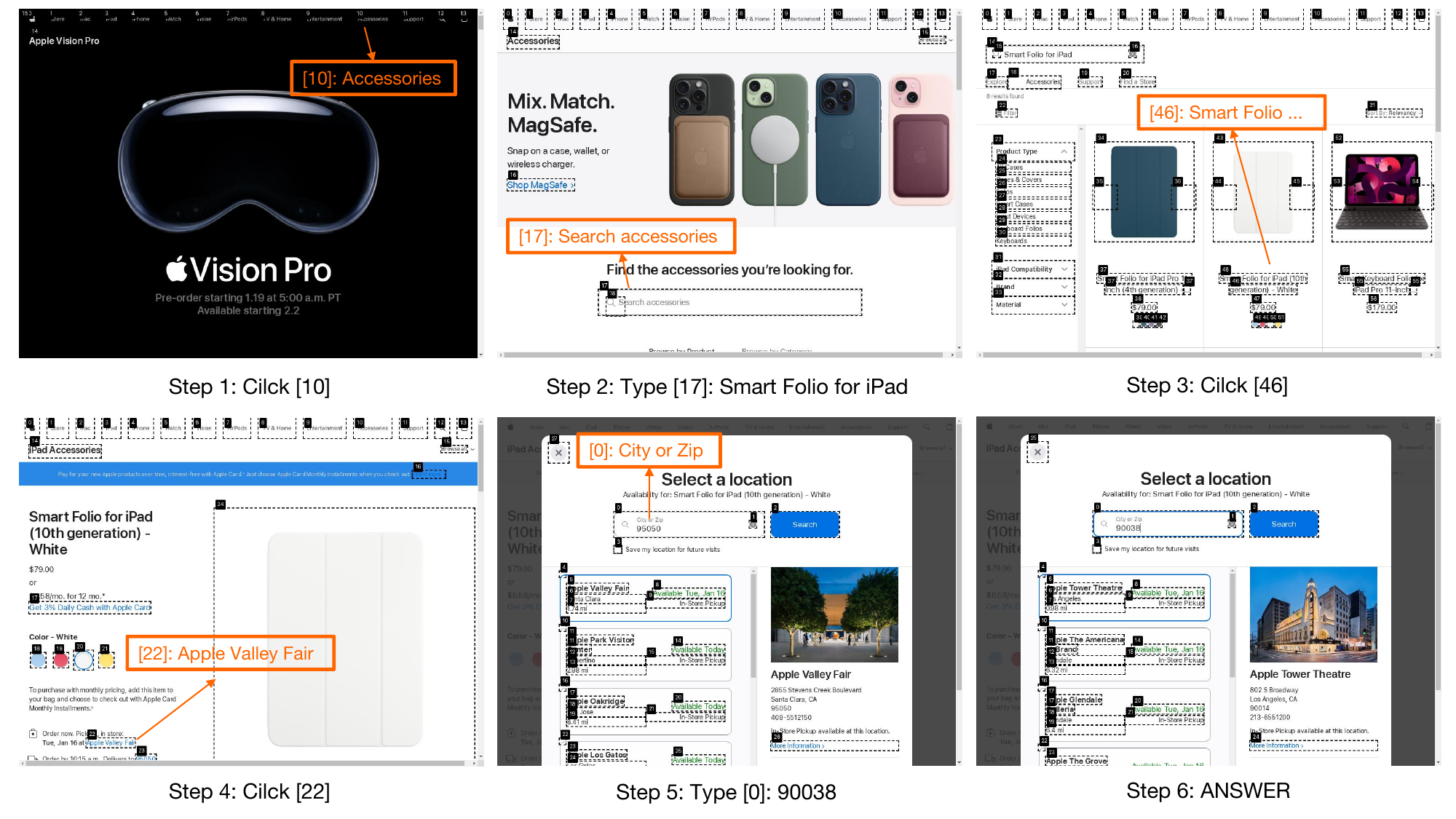}
\caption{Screenshots of a complete trajectory of online web browsing. Given the task: `Search Apple for the accessory Smart Folio for iPad and check the closest pickup availability next to zip code 90038.' The agent interacts with the Apple website and obtains the answer: `Apple Tower Theatre.'}
\label{fig:episode}
\vspace{-0.1in}
\end{figure*}

\section{Experiment}
\paragraph{Dataset and Metrics}
Besides the benchmark introduced in Section \ref{sec:benchmark}, we also evaluated on 90 web browsing tasks (Level 1 and Level 2) from the GAIA dataset \citep{mialon2023gaia}, which also come with golden responses. Since GAIA does not provide specific websites, we instruct the agent to start with Google Search. We further evaluated our agent on the 50 tasks used in SeeAct agent's online evaluation \cite{zheng2024gpt} and compared with their results. Following WebArena \citep{zhou2023webarena}, the primary evaluation metric we adopt is the \textbf{Task Success Rate}, measuring the successful completion of tasks without considering whether the steps are optimal. 

\paragraph{Experimental Details}
We employ GPT-4 Turbo with vision (gpt-4-vision-preview) as the backbone model of our agent, which showcases strong semantic and visual understanding capabilities equivalent to GPT-4V \citep{openai2023gpt4}. Furthermore, we incorporate two additional backbone models, namely Claude 3 Opus \citep{anthropic2024introducing} and GPT-4o (GPT-4 Omni, \citealp{openai2024hello}), to augment the diversity of our research. We include the GPT-4 (All Tools) for baselines, which integrates vision, web browsing, code analysis, and various plugins in one model. Additionally, we consider a text-only baseline where the agent only receives the websites' accessibility tree as input to prediction actions.  
We used a fixed browser window size of 1024 * 768 pixels for our environment, ensuring a consistent size for the screenshots in our observations. We set the temperature to 1 during generation and allow the agent to explore at most 15 steps. 

\subsection{Evaluation Methods}
We adopt human evaluation as our main evaluation metric since most of the questions in our benchmark have open-ended answers. In particular, we provide the human evaluators with the complete trajectories of the agent's interaction with the web (all screenshots and all actions), and ask them to provide a binary judgment of whether the agent successfully completed the task. 
For a subset of 300 tasks, we invite three annotators to judge each trajectory to understand the agreement among human annotators. 

Even though human evaluations are accurate, they are often not scalable.
Hence, we want to see if leveraging an LMM for automatic evaluation is feasible. To this end, we propose to use GPT-4V as an auto-evaluator that emulates the behavior of human evaluators to evaluate the navigation trajectories of WebVoyager. In particular, we provide the task, the responses from WebVoyager, and the last $k$ screenshots to the evaluator and ask it to judge whether the agent has completed the task, where $k$ is a hyper-parameter. The prompt of the GPT-4V evaluator is shown in Appendix \ref{appendix:auto_eval}.



\begin{table*}[t]
\centering
\setlength{\tabcolsep}{1.6mm}{
\scalebox{0.72}{\begin{tabular}{@{}lcccccccc@{}}
\toprule \toprule
& Allrecipes & Amazon & Apple & ArXiv & GitHub & Booking & ESPN & Coursera \\ \midrule
GPT-4 (All Tools) & \ \ 11.1\% & \ \ 17.1\% & \ \ 44.2\% & \ \ 14.0\% & \ \ 48.8\% & \ \ 22.7\%  & \ \ 31.8\% & \ \ 31.0\% \\
WebVoyager$_{\text{Text-only}}$ & \ \ \bf 55.6\% & \ \ 31.7\% & \ \ 34.9\%  & \ \ 32.6\% & \ \ 61.0\% & \ \ 2.3\% & \ \ 36.4\% & \ \ 23.8\% \\
WebVoyager  & \ \ 53.3\% & \ \ \bf 58.5\%  & \ \ \bf 65.1\%  & \ \ \bf 51.2\% & \ \ \bf 63.4\% & \ \ \bf 43.2\%  & \ \ \bf 38.6\% & \ \ \bf 73.8\% \\ 

\textit{WebVoyager$_{\text{Text-only}}$}$^*$ & \ \ 57.8\%$_{\pm 0.0\%}$ & \ \ 43.1\%$_{\pm 1.4\%}$ & \ \ 36.4\%$_{\pm 3.5\%}$ & \ \ 50.4\%$_{\pm 1.4\%}$ & \ \ 63.4\%$_{\pm 2.5\%}$ & \ \ 2.3\%$_{\pm 0.0\%}$ & \ \ 38.6\%$_{\pm 2.3\%}$ & \ \ 24.6\%$_{\pm 1.4\%}$ \\

\textit{WebVoyager}$^*$ & \ \ 51.1\%$_{\pm 2.2\%}$ & \ \  52.9\%$_{\pm 1.4\%}$  & \ \ 62.8\%$_{\pm 2.3\%}$  & \ \ 52.0\%$_{\pm 1.3\%}$ & \ \ 59.3\%$_{\pm 3.7\%}$ & \ \ 32.6\%$_{\pm 2.7\%}$  & \ \  47.0\%$_{\pm 1.3\%}$  & \ \  57.9\%$_{\pm 2.7\%}$   \\ 

\textit{WebVoyager$_{\text{Claude}}$}$^*$ & \ \ 45.9\%$_{\pm 3.4\%}$ & \ \  58.6\%$_{\pm 4.2\%}$  & \ \ 58.1\%$_{\pm 4.0\%}$  & \ \ 55.0\%$_{\pm 7.0\%}$ & \ \ 56.9\%$_{\pm 1.4\%}$ & \ \ 19.0\%$_{\pm 1.3\%}$  & \ \  46.2\%$_{\pm 1.3\%}$  & \ \  68.2\%$_{\pm 1.3\%}$   \\

\textit{WebVoyager$_{\text{GPT-4o}}$}$^*$ & \ \ 56.3\%$_{\pm 1.3\%}$ & \ \  53.7\%$_{\pm 2.5\%}$  & \ \ 56.6\%$_{\pm 1.3\%}$  & \ \ 60.5\%$_{\pm 0.0\%}$ & \ \ 57.7\%$_{\pm 3.7\%}$ & \ \ 43.9\%$_{\pm 3.5\%}$  & \ \  44.0\%$_{\pm 2.7\%}$  & \ \  65.1\%$_{\pm 2.8\%}$   \\

\midrule \midrule
& Cambridge & BBC & Google & Google & Google & \multirow{2}{*}{Huggingface} & Wolfram & \multirow{2}{*}{Overall}  \\ 
& Dictionary & News & Flights & Map & Search & & Alpha & \\ 
\midrule 
GPT-4 (All Tools)  & \ \ 25.6\% & \ \ 9.5\%  & \ \ 2.4\%  & \ \ 53.7\% & \ \ 60.5\% & \ \ 37.2\%  & \ \ 52.2\% & \ \ 30.8\%  \\
WebVoyager$_{\text{Text-only}}$ & \ \ 62.8\% & \ \ 45.2\%  & \ \ 7.1\% & \ \  61.0\% & \ \ 67.4\% & \ \ 20.9\% & \ \ 58.7\% & \ \ 40.1\%  \\
WebVoyager  & \ \  \bf 65.1\% & \ \ \bf 61.9\%  & \ \ \bf 59.5\%  & \ \ \bf 70.7\% & \ \ \bf 76.7\% & \ \ \bf 44.2\%  & \ \ \bf 63.0\% & \ \ \bf 59.1\%  \\ 
\textit{WebVoyager$_{\text{Text-only}}$}$^*$ & \ \ 66.7\%$_{\pm 3.6\%}$ & \ \ 45.2\%$_{\pm 2.4\%}$ & \ \ 7.1\%$_{\pm 0.0\%}$ & \ \ 62.6\%$_{\pm 2.8\%}$ & \ \ 75.2\%$_{\pm 1.3\%}$ & \ \ 31.0\%$_{\pm 1.4\%}$ & \ \ 60.2\%$_{\pm 1.3\%}$ & \ \ 44.3\%$_{\pm 0.6\%}$ \\ 
\textit{WebVoyager}$^*$ & \ \ 71.3\%$_{\pm 1.3\%}$ & \ \  60.3\%$_{\pm 2.8\%}$  & \ \ 51.6\%$_{\pm 1.4\%}$  & \ \ 64.3\%$_{\pm 2.8\%}$ & \ \ 77.5\%$_{\pm 2.7\%}$ & \ \ 55.8\%$_{\pm 2.3\%}$  & \ \  60.9\%$_{\pm 2.2\%}$  & \ \  57.1\%$_{\pm 0.2\%}$  \\ 

\textit{WebVoyager$_{\text{Claude}}$}$^*$ & \ \ 71.3\%$_{\pm 3.6\%}$ & \ \  66.7\%$_{\pm 4.8\%}$  & \ \ 15.1\%$_{\pm 5.5\%}$  & \ \ 55.3\%$_{\pm 1.4\%}$ & \ \ 72.9\%$_{\pm 1.3\%}$ & \ \ 53.5\%$_{\pm 4.7\%}$  & \ \  51.5\%$_{\pm 5.4\%}$  & \ \  52.8\%$_{\pm 1.4\%}$  \\ 

\textit{WebVoyager$_{\text{GPT-4o}}$}$^*$ & \ \ 82.2\%$_{\pm 1.3\%}$ & \ \  54.8\%$_{\pm 2.4\%}$  & \ \ 28.6\%$_{\pm 0.0\%}$  & \ \ 56.9\%$_{\pm 2.8\%}$ & \ \ 63.6\%$_{\pm 1.3\%}$ & \ \ 42.6\%$_{\pm 3.6\%}$  & \ \  65.2\%$_{\pm 2.2\%}$  & \ \  55.5\%$_{\pm 0.8\%}$  \\
\bottomrule \bottomrule
\end{tabular}}}
\caption{
The main result for WebVoyager.
Each website contains 40 to 45 tasks, and we report the Task Success Rate in the table.
We show the results of GPT-4 (All Tools), WebVoyager$_{\text{Text-only}}$ using the accessibility tree, and WebVoyager by comparing with human expert labels. \textit{WebVoyager}$^*$, \textit{WebVoyager$_{\text{Text-only}}$}$^*$, \textit{WebVoyager$_{\text{Claude}}$}$^*$ and \textit{WebVoyager$_{\text{GPT-4o}}$}$^*$  are results evaluated by GPT-4V (full trajectory, kappa = 0.70). For each automatic evaluation, we run GPT-4V evaluator three times to calculate the performance mean and standard deviation.} 
\label{tab:noise}
\end{table*}

\begin{table}[t]
\small
\vspace{-0.1in}
\centering 
\begin{tabular}{@{}lccc@{}}
\toprule
\multirow{2}{*}{} & \multirow{2}{*}{Success Rate} & \multicolumn{2}{c}{Consistency} \\ \cmidrule(l){3-4} 
                  &                                       & Agreement        & $\kappa$        \\ \midrule
k=1               & 47.7\%                                & 75.3\%           & 0.51         \\
k=2               & 55.3\%                                & 79.7\%           & 0.59         \\
k=3               & 54.3\%                                & 81.3\%           & 0.62         \\
Full   & 58.3\%                                & 85.3\%           & 0.70         \\ \bottomrule
\end{tabular}
\caption{Consistency between GPT-4V and Human. Success Rate is the overall success rate of all tasks given by GPT-4V. Based on the annotations given by GPT-4V and Human (after alignment), we compute Agreement, i.e., the label overlap, and the Kappa values.}
\label{tab:consistency}
\end{table}

\subsection{Result}
Figure \ref{fig:episode} presents an example that demonstrates how the agent interacts with the Apple website step by step in an online fashion to complete a task. In the final screenshot, the Agent acquires the desired information, then selects the    ``ANSWER'' action to respond and conclude the navigation. 
Additional examples are provided in the Appendix \ref{sec:additional_traj}. 

\begin{figure}[t!]
\centering
\includegraphics[width=0.8\linewidth]{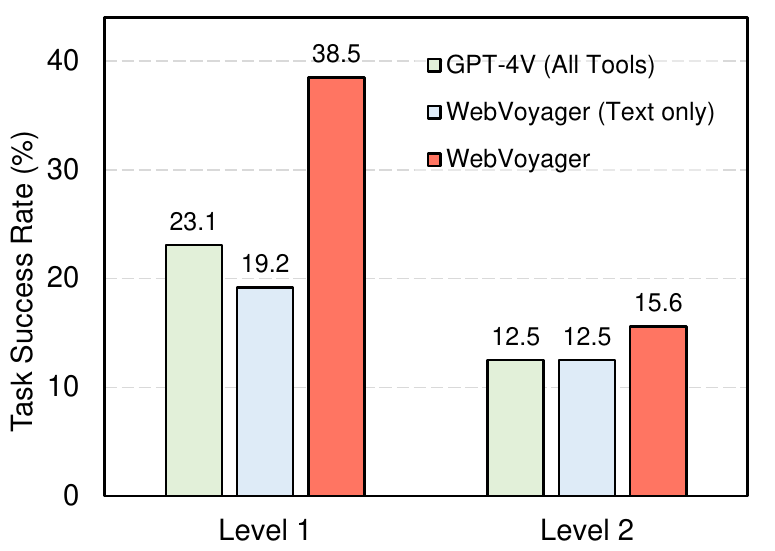}
\caption{Success Rate for GAIA Level 1 and Level 2. }
\vspace{-0.1in}
\label{fig:success_rate_gaia}
\end{figure}

We present the results for our dataset and the extracted GAIA web tasks in Table \ref{tab:noise} and Figure \ref{fig:success_rate_gaia}. 
WebVoyager outperforms text-only and GPT-4 (All Tools) baselines by large margins in most website tasks, while it is slightly lower than Text-only on Allrecipes and similar to Text-only on Github, ESPN, Cambridge Dictionary and Wolfram Alpha. This is primarily because these websites are more text-heavy than others. Since WebVoyager mostly relies on web screenshots for decision-making, dense text might not be easily recognizable from the image. We think extracting such text from the HTML to augment the input could be a potential solution to this problem, suggesting a direction for future work. 
In Figure \ref{fig:success_rate_gaia}, WebVoyager also achieves much stronger performance than both baselines. Finally, WebVoyager has a success rate of 30\% on the SeeAct online test set whereas the best SeeAct autonomous agent has 26\%, showing the efficacy of our proposed agent.

We report the Agreement (the ratio of overlap) and Kappa ($\kappa$; \citealt{cohen1960coefficient}) between consolidated human labels\footnote{the Fleiss's Kappa \citep{fleiss1971measuring} of human annotators before any discussion is 0.7, which is substantial agreement. } and GPT-4V's judgments on the subset of 300 tasks in Table \ref{tab:consistency}. Here,
$k$ denotes the number of screenshots provided to GPT-4V, with ``full'' implying the full trajectory. GPT-4V's agreement with human annotators gradually improves as it receives more information, and its final Kappa score also reaches 0.7, which is on par with the agreement among human annotators. The consistency between GPT-4V and humans suggests that GPT-4V is a promising automatic evaluator for multi-modal web agents. Accordingly, we report the automatic evaluation results based on GPT-4V in Table \ref{tab:noise}. The automatic evaluation results of three backbone models, GPT-4V, Claude 3 Opus, and GPT-4o, are relatively close, and their performance is significantly better than the Text-only setting (with GPT-4 as the backbone). 
However, there is a performance decline for both Claude and GPT-4o on Google Flights. Upon reviewing the trajectories, it is observed that GPT-4o falls into a cognitive bias, where it fails to correctly select the `one way' option for one-way trip tasks, mistakenly assuming that only the departure date needs to be entered. On the other hand, Claude-3-Opus encounters difficulties in correctly interacting with web elements while inputting basic flight information. Modifying the system prompt for GPT-4o or Claude may potentially improve the performance. 

\begin{table}[t]
\small
\vspace{-0.1in}
\centering 
\begin{tabular}{@{}cccc@{}}
\toprule
\multirow{2}{*}{\begin{tabular}[c]{@{}c@{}}WebVoyager \\ Backbone\end{tabular}} & \multicolumn{3}{c}{Evaluator}   \\ \cmidrule(l){2-4}  & GPT-4V & Claude-3-Opus & GPT-4o \\ \midrule
GPT-4V     & 57.1   & 55.1          & 63.0     \\
Claude-3-Opus  & 52.8   & 61.6          & 55.4   \\
GPT-4o    & 55.5   &  54.9             & 64.1   \\ \bottomrule
\end{tabular}
\caption{Overall Task Success Rate of WebVoyager using automatic evaluation. We employ GPT-4V, Claude-3-Opus, and GPT-4o as backbones for WebVoyager and run all tasks, followed by automatic evaluation using these three models.}
\label{tab:auto_eval_mat}
\end{table}

Besides, we conduct the Claude-3-Opus based evaluation and the GPT-4o based evaluation. When provided with the full trajectory, the Claude-3-Opus achieves a kappa value of 0.6 with humans, indicating that it is less reliable than the GPT-4V. And the kappa value between GPT-4o and humans is 0.72, slightly higher than that of GPT-4V. Table \ref{tab:auto_eval_mat} illustrates the Task Success Rate when using GPT-4V, Claude-3-Opus, and GPT-4o as backbones and auto evaluators. We observe that GPT-4o exhibits a more lenient attitude towards task performance results, while GPT-4V tends to be relatively strict. However, both models agree that Claude-3-Opus performs the worst. Claude-3-Opus, on the other hand, demonstrates a clear preference for its own results, believing that GPT-4V and GPT-4o are similar but considers itself to have the best performance. GPT-4o and GPT-4V also exhibit a certain bias towards their own results, with each considering itself to be superior to the other.

\subsection{Discussions}
\textbf{Direct interaction with the websites is necessary} From our experience of using GPT-4 (All Tools), the primary limitation of GPT-4 (All Tools)'s performance is rooted in its reliance on Bing search for web browsing, predominantly depending on web pages fetched by Bing. It cannot directly access certain websites (such as Apple, Amazon, BBC News, etc.) for searching, clicking, or utilizing their sorting functions. This greatly limits the agent's ability to complete certain types of tasks. 

\noindent \textbf{Both text and vision are necessary for generalist web agents}. As discussed earlier, WebVoyager struggles with text-heavy websites. On the other hand,  we observe that the text-only agent demonstrates significantly poorer performance on websites with complex visual elements, such as Booking and Flights, which require interactions with calendars and other intricate components. In these scenarios, the textual input such as the accessibility tree becomes highly complex and verbose, making it far less intuitive than using screenshots. Hence it's necessary to incorporate both modalities of inputs when building the general purpose agents.

\begin{figure}[t!]
\centering

\includegraphics[width=0.9\linewidth]{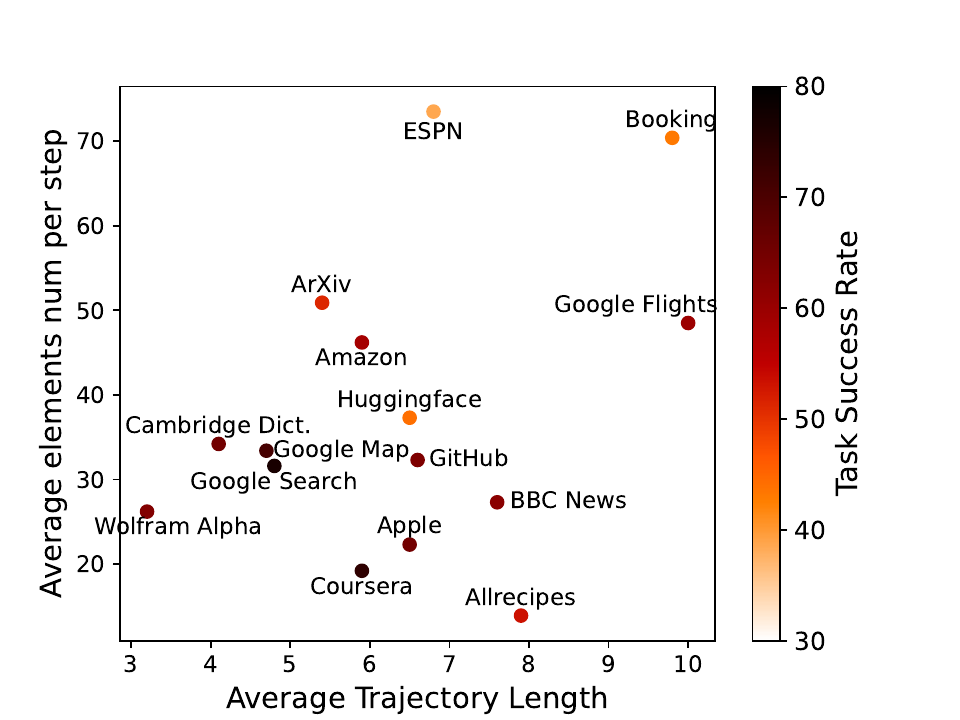}
\caption{Factors related to task success rate. We show the average number of elements per page and the average trajectory length for each website type. The darker colors indicate a higher task success rate.}
\label{fig:page_complexity}
\vspace{-0.1in}
\end{figure}

\noindent \textbf{Websites with more interactable elements are more challenging for agents.} We also calculate the average trajectory length of tasks completed within the maximum number of iterations, as well as the average number of interactive web elements present on the webpage screenshots.  Figure \ref{fig:page_complexity} illustrates their relationship with the Task Success Rate. We posit that the average trajectory length serves as a measure of a task's complexity to some extent, while the average number of numerical labels related to the decision-making process reflects the complexity of a webpage.  Intuitively, websites depicted in the lower-left corner of Figure \ref{fig:page_complexity}, characterized by relatively simple webpages and shorter trajectory lengths, are expected to exhibit higher Task Success Rates. As observed in Figure \ref{fig:page_complexity}, the results largely align with this intuition. 

\noindent \textbf{Why not use Open Source models.} There are a few critical limitations of the existing open-sourced LMMs that prevent us from exploring other alternatives. Specifically, the web navigation task requires the agent to process fine-grained details from web page screenshots to make decisions, hence high-resolution is required to preserve the information from the web. However, most open-sourced LMMs such as  LLaVA \citep{liu2024visual} reduce the image resolution to 224x224 or 336x336, which makes text with smaller fonts unrecognizable, hence they are unsuitable for web navigation tasks. Moreover, models like LLaVA have a max context length of 4096. In our case, the agent needs to handle trajectories as long as 15 steps, and it requires approximately 7000+ tokens, which does not fit in those models' context size.

\subsection{Error Analysis}
In this section, we discuss and summarize the primary issues encountered by WebVoyager in the task completion process. These challenges will serve as critical entry points for future enhancements of the Task Success Rate and for devising strategies to obtain an Optimal Trajectory. We sampled 300 tasks from our benchmark and manually labeled the error category for each failed case, we show the distribution of errors in table \ref{tab:error}. 
In Appendix \ref{sec:error_eg}, we also provide specific examples for each issue. 

\paragraph{Navigation Stuck} 
The most common failure is running out of steps before completing the task. There are three failure scenarios: 
1) When the agent's search query is not precise and explicit enough, it will be overwhelmed by irrelevant search results. The agent may prefer to browse different results or wait for incorrect outcomes rather than correct its previous action; 2) When the scroll-able area is 
very small,
the agent might not be able to locate the correct scrolling area and repeatedly request the execution of useless scrolling actions; 3) Sometimes in the middle of the page, the agent has trouble deciding whether to scroll up or down.
The agent also tends to repeat its previous mistakes due to the input clipping, as mentioned in section \ref{subsec:planning_method}. 
These meaningless or repetitive actions may hinder the completion of the task.

\paragraph{Visual Grounding Issue} The visual grounding ability of our agent still has a large room for improvement. We observe the following issues: 1) The agent cannot interpret less frequently observed patterns, such as misidentifying characters representing the pronunciations or math formulas; 2) The agent doesn't recognize the subtle difference between two observations and thinks the execution has failed;  3) The agent selects the wrong element for action execution due to proximity. 
For example, the model sometimes confuses adjacent elements and misinterprets numbers on a calendar as numerical labels. Sometimes textual information plays a significant role, offering valuable cues and assisting in distinguishing between overly dense web elements. We find that incorporating the text content included in Web Elements can alleviate these problems to some extent. However, a stronger visual encoder or additional text inputs might be needed.

\begin{table}[t]
\small
\centering 
\begin{tabular}{@{}lll@{}}
\toprule
\begin{tabular}[c]{@{}l@{}}Main reasons for \\ Failure\end{tabular} & Ratio  \\ \midrule
Navigation Stuck &   44.4\%     \\
Visual Grounding Issue    &  24.8\%      \\
Hallucination&   21.8\%     \\
Prompt Misalignment     &    9.0\%     \\ \bottomrule
\end{tabular}
\caption{Distribution of main failure reasons.}
\vspace{-0.1in}
\label{tab:error}
\end{table}

\paragraph{Hallucination} 
Agents sometimes produce seemingly correct answers, which may require checking carefully to identify errors. We mainly see the following two scenarios: 1) The agent may overlook certain task requirements and settle for an answer that is only partially correct. For instance, when asked for the cheapest product, the agent might respond with a cheap product visible in a screenshot, neglecting the need to sort the options first.
2) The agent might execute a seemingly correct action without raising any errors, which deviate it from the correct reasoning path. 
For example, inputting content to the wrong text box when there are many text boxes on the webpage is still valid, yet it would guide the agent to obtain a wrong answer.

\paragraph{Prompt Misalignment} 
Understanding and following complex prompts, as illustrated in Figure \ref{fig:prompt}, often poses significant challenges. Moreover, longer trajectories may result in excessively lengthy contexts, hindering effective instruction following. Although many of the errors in Navigation Stuck and Hallucination categories can also be attributed to prompt design, we use Prompt Misalignment to categorize the following situations: 1) the agent fails to generate outputs that can be parsed into executable actions, e.g. providing only the `Thought' without the corresponding `Action'; 2) Prematurely terminating the process using the ANSWER action, though agent knows that the task is not yet complete (explicitly mentioned in its answer). 

\section{Conclusion}

We introduce WebVoyager, an innovative web agent powered by large multimodal models (LMM) that can complete real-world web tasks end-to-end by interacting with websites. We have shown through evaluations that WebVoyager outperforms several baselines by leveraging both visual and textual signals. 
We also propose an automatic evaluation protocol by leveraging GPT-4V as the evaluator for online agents. 
Our work demonstrates the promise of using advanced LMM capabilities in building intelligent web agents. We hope WebVoyager provides a strong foundation for future research toward building more versatile and capable web assistants.

\section*{Limitations}
We recognize the following limitations of our work.
First, we haven't supported all possible actions in our environment compared to actions a human user might take when browsing the web. e.g. the Drag action on web pages. Supporting such an action is challenging since the degree of Drag is not a finite set. We may allow it to choose the pixel values to be dragged if the Visual Grounding capabilities of LMMs are further enhanced. Second, our agent currently can only analyze basic file formats (such as text files and PDF files) and doesn't support all file formats, especially videos. Enhancing support for additional file formats is a crucial step in the development of web agents and we leave it for future work. 

Regarding the potential risks of our work, we believe that it requires a substantial amount of safety checks before deploying web agents like WebVoyagar into real-world applications, as the agent might unintentionally download malicious content from unauthorized websites, or input private/confidential information on public websites. Also the agent might send fake requests to website servers or generate fake user activities, which might be harmful to website owners. Therefore it's necessary to take extra caution when using and testing our agent. 

\section*{Ethics Statement}
Our experiments have been designed to operate within strict ethical guidelines. Specifically, we restrict our web agent to perform only non-login tasks. This approach is in full compliance with the terms of service and user agreements of the websites our agent interacts with. Furthermore, we closely monitor the agent's activities during its online evaluations. This monitoring is designed to identify and prevent any actions that could lead to potentially harmful consequences. By taking these precautions, we ensure that our research does not cross ethical boundaries or cause unintended harm.

Additionally, all task queries for evaluation undergo thorough manual inspection to ensure they are harmless and ethically sound. This manual inspection process is aimed at ensuring that every query is harmless and does not promote or propagate harmful content or actions. Our work aims to enhance user experience and accessibility while mitigating potential negative societal impacts. By proactively addressing ethics concerns, we dedicate to conducting research that benefits society while upholding high ethical standards.


\bibliography{custom}
\bibliographystyle{acl_natbib}

\clearpage
\appendix

\section{Prompt for WebVoyager}
\label{sec:appendixA}

\begin{figure*}[!h]
\centering
\includegraphics[width=0.97\linewidth]{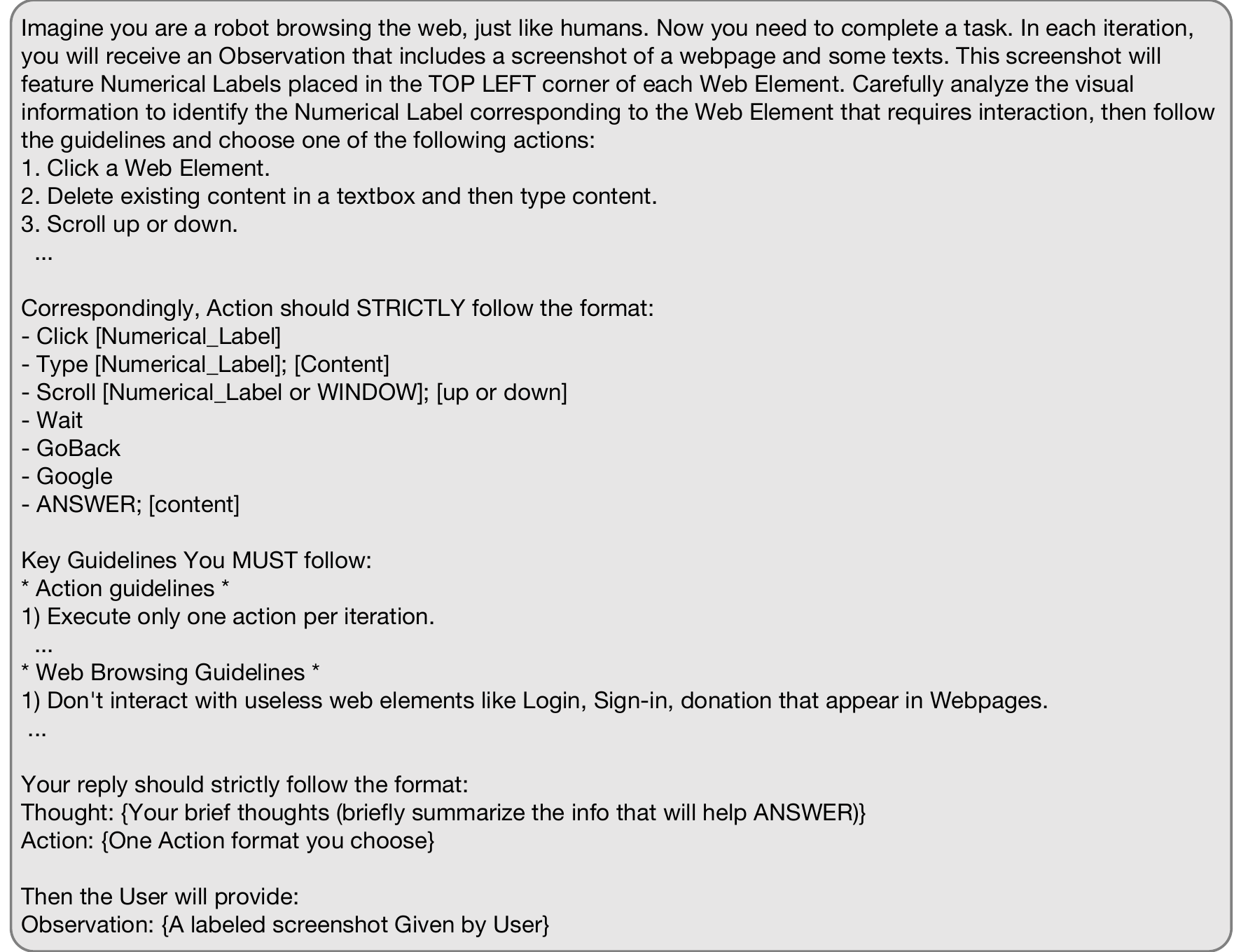}
\caption{System Prompt for WebVoyager. We instruct agents to perform web navigation, along with specific browsing actions and action formats. To enhance efficiency and accuracy, we can incorporate additional general into the prompts. These guidelines should be generic and not about a specific website to ensure generalizability.}
\label{fig:prompt}
\end{figure*}

The System Prompt for WebVoyager is shown in Figure \ref{fig:prompt}. The Prompt's guidelines hold potential for optimization and should be generalist rather than website-specific in design. Incorporating specific issues from websites into the system prompt may compromise the agent's universality.

\section{Prompt for Auto Evaluation}\label{appendix:auto_eval}

\begin{figure*}[h!]
\centering
\includegraphics[width=0.98\linewidth]{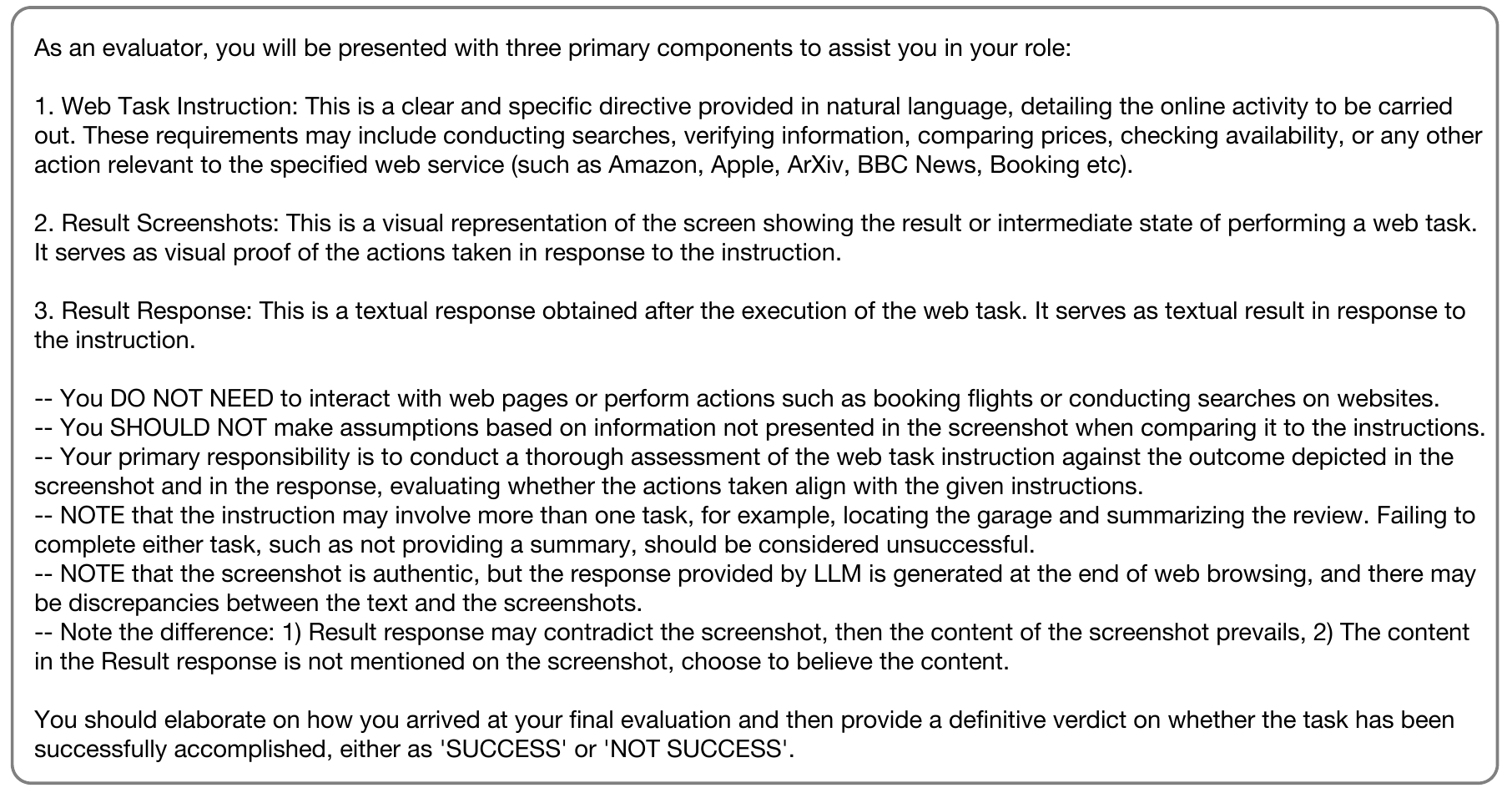}
\caption{System Prompt for Auto Evaluation by GPT-4V.}
\label{fig:prompt_auto}
\end{figure*}

Figure \ref{fig:prompt_auto} demonstrates using GPT-4V as an evaluator for web tasks, involving web task instruction, screenshots in the trajectory, and WebVoyager responses. We require GPT-4V to mark success or not success. The temperature is set to 0 to reduce randomness during evaluation.

\section{Action Space}\label{sec:action_space}
We detail the interaction actions that WebVoyager employs to navigate and operate within web environments. These actions are fundamental to how the agent interacts with web pages, retrieves information, and performs specific tasks as part of its operational protocol. The actions range from basic web navigation to more complex operations, tailored to efficiently gather data and respond to queries. Each action is designed with a specific format for easy identification and execution. 

\begin{itemize}
    \item Click. This action involves clicking on an element within a webpage, typically a link or a button. 
    If clicking a link results in the download of a PDF file, we automatically parse its content using the OpenAI Assistant API\footnote{https://platform.openai.com/docs/assistants/overview} and incorporate it into the Observation. Action Format: \texttt{Click [Numerical\_Label]}.
    \item Input. This is a composite action that involves selecting a text box, deleting any existing content within it, and then inputting new content. 
    To minimize interaction frequency, an automatic ENTER key press follows the input completion. Action Format:  \texttt{Type [Numerical\_Label]; [Content]}.
    \item Scroll. Scrolling is a common operation for browsing webpages, usually involving the vertical movement of the entire page. 
    However, there are instances where only a specific section within the webpage is scrollable. In such cases, we expect the Agent to select an element within the scrollable area and navigate to that particular region for scrolling. Action Format: \texttt{Scroll [Numerical\_Label or WINDOW]; [up or down]}.
    \item Wait. Action execution requires time, and this action is often used to wait for web pages to load. 
    Action Format: \texttt{Wait}. 
    \item Back. This action is used to return to the previous page. 
    We consider the forward action unnecessary because it can be achieved by repeating previous actions. Action Format: \texttt{GoBack}.
    \item Jump to Search Engine. There are often situations where agents get stuck at a certain website, without finding an answer. This action enables the agent to jump to a search engine and start anew. 
    In this work, we just adopt Google Search. Action Format: \texttt{Google}. 
    \item Answer. Once all questions in the task are resolved, this action concludes the iteration and provides an answer in line with the task requirements. 
    Action Format: \texttt{ANSWER; [Content]}.

\end{itemize}

\section{Additional Trajectories}\label{sec:additional_traj}
\begin{figure*}[t!]
\centering
\includegraphics[width=1.0\linewidth]{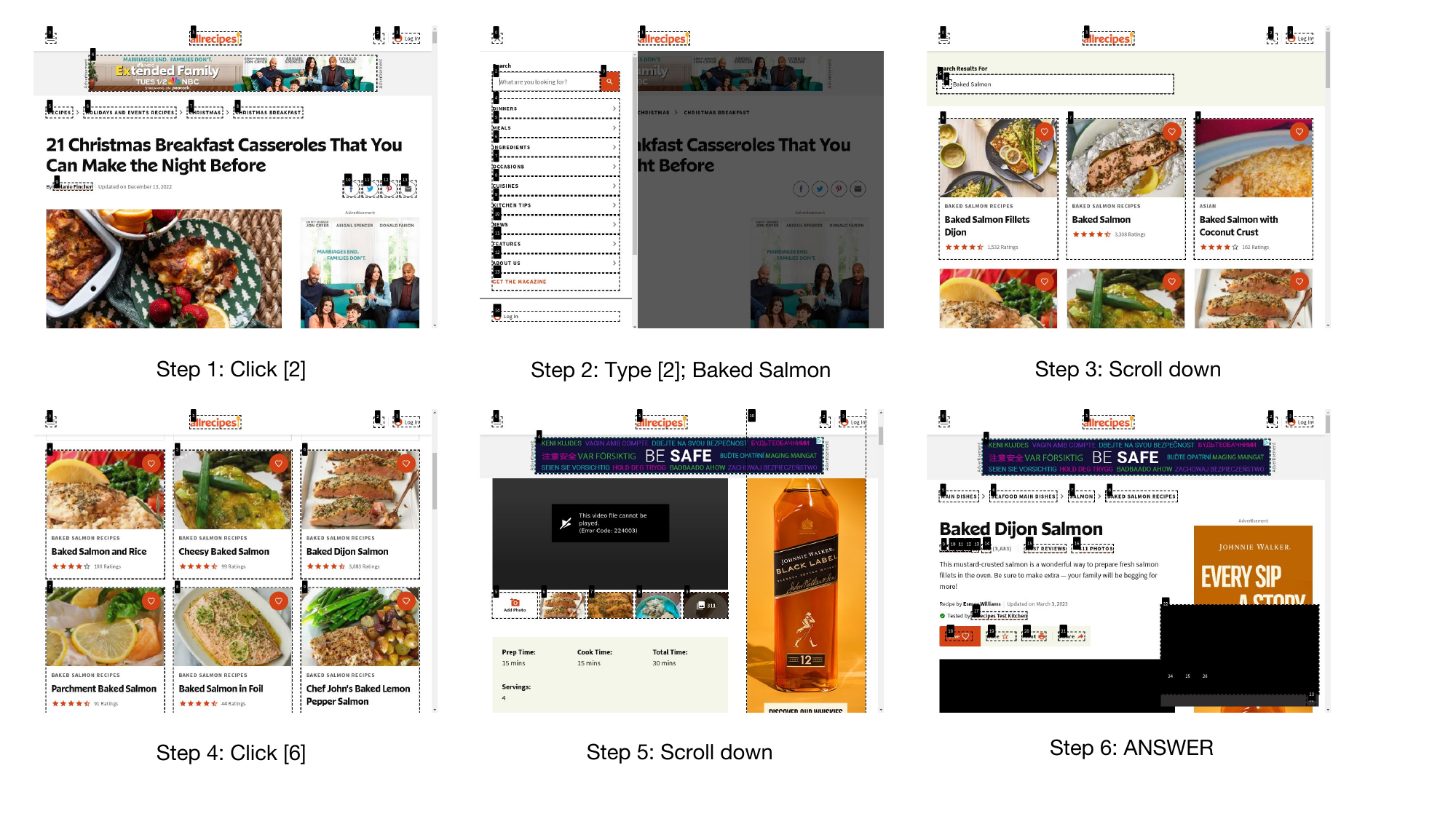}
\caption{Screenshots of a complete trajectory of browsing Allrecipes. Given the task: ``Find a recipe for Baked Salmon that takes less than 30 minutes to prepare and has at least a 4-star rating based on user reviews.'' The agent interacts with the Allrecipes website and obtains the answer: ``The "Baked Dijon Salmon" recipe meets the user's criteria, with a 4.6-star rating and a preparation time of 15 minutes.''}
\label{fig:episode_allrecipes}
\end{figure*}

\begin{figure*}[t!]
\centering
\includegraphics[width=1.0\linewidth]{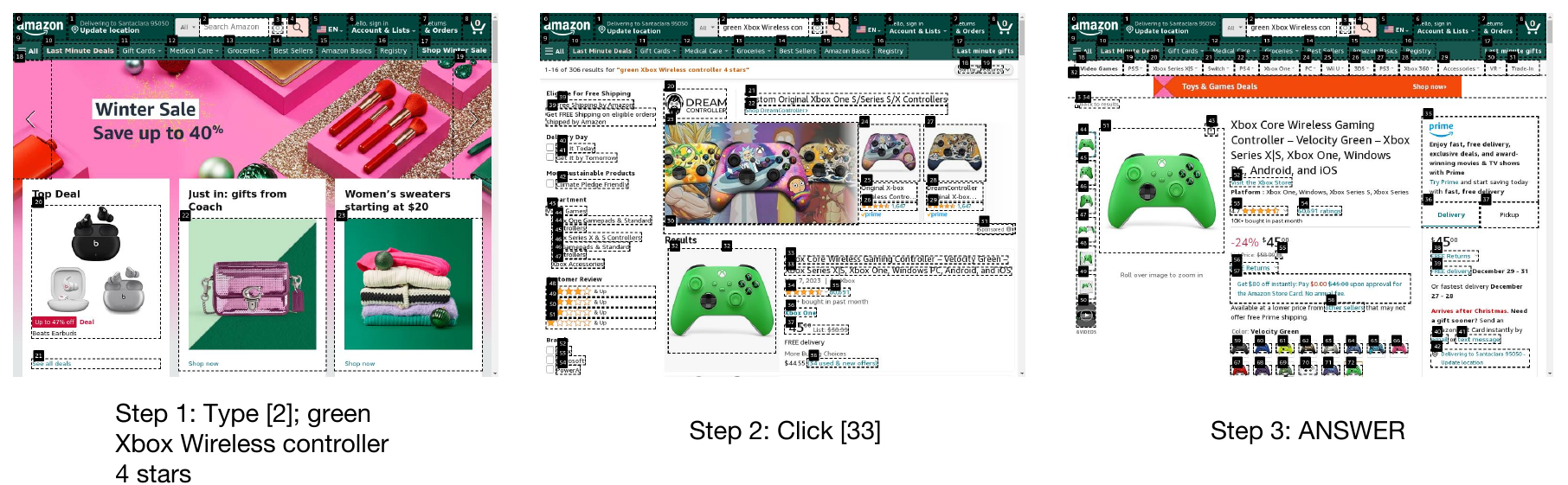}
\caption{Screenshots of a complete trajectory of browsing Amazon. Given the task: ``Search for an Xbox Wireless controller with green color and rated above 4 stars.'' The agent interacts with the Amazon website and obtains the answer: ``The green Xbox Wireless controller ("Xbox Core Wireless Gaming Controller – Velocity Green") rated above 4 stars has been found on Amazon with a rating of 4.7 out of 5 stars.''}
\label{fig:episode_amazon}
\end{figure*}

\begin{figure*}[t!]
\centering
\includegraphics[width=1.0\linewidth]{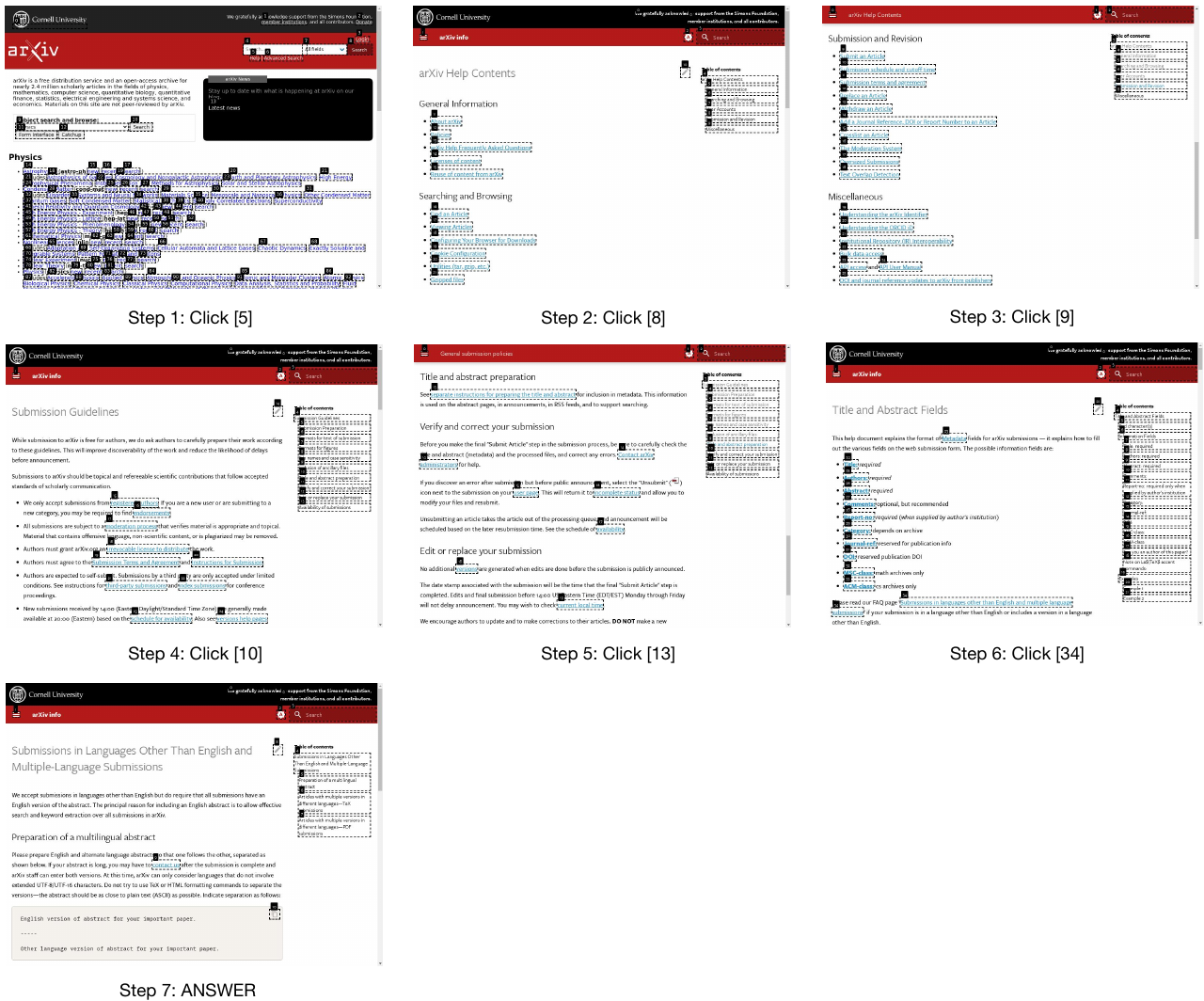}
\caption{Screenshots of a complete trajectory of browsing ArXiv. Given the task: ``For non-English submissions, do I need to provide a multi-language abstract, if needed, answer the separator between the multiple abstracts.'' The agent interacts with the ArXiv website and obtains the answer: ``Line ("\textrm{-{}-{}-{}-{}-}")''}
\label{fig:episode_arxiv}
\end{figure*}

\begin{figure*}[t!]
\centering
\includegraphics[width=1.0\linewidth]{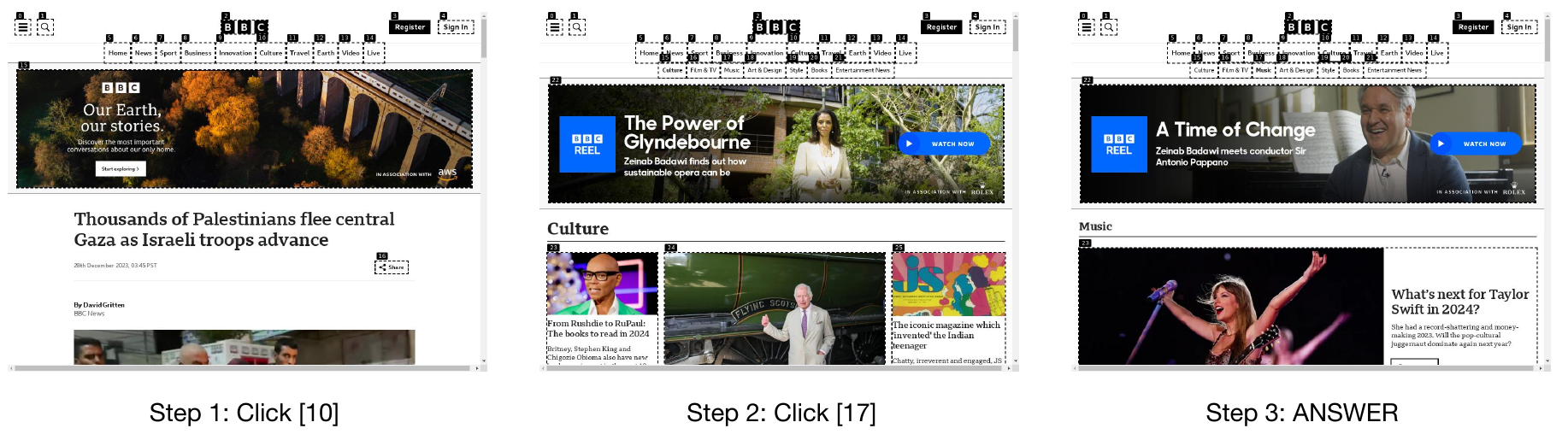}
\caption{Screenshots of a complete trajectory of browsing BBC News. Given the task: ``Find out which musician made the headlines in Music News.'' The agent interacts with the BBC News website and obtains the answer: ``The musician who made the headlines in Music News is Taylor Swift.''}
\label{fig:episode_bbc}
\end{figure*}

\begin{figure*}[t!]
\centering
\includegraphics[width=1.0\linewidth]{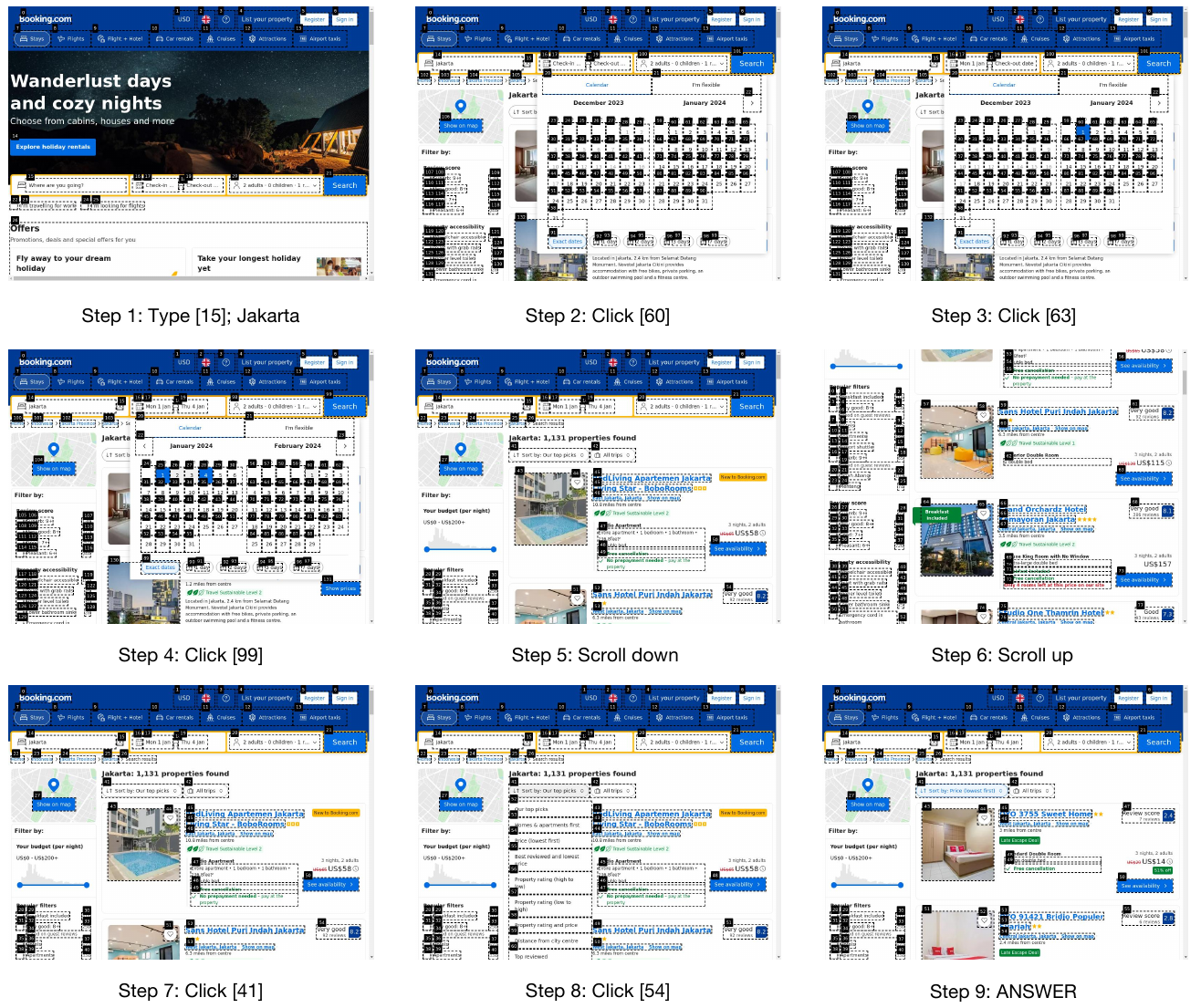}
\caption{Screenshots of a complete trajectory of browsing Booking. Given the task: ``Find the cheapest available hotel room for a three-night stay from 1st Jan in Jakarta. The room is for 2 adults, just answer the cheapest hotel room and the price.'' The agent interacts with the Booking website and obtains the answer: ``The cheapest hotel room is at OYO 3755 Sweet Home for US\$14 for a three-night stay.''}
\label{fig:episode_booking}
\end{figure*}

\begin{figure*}[t!]
\centering
\includegraphics[width=1.0\linewidth]{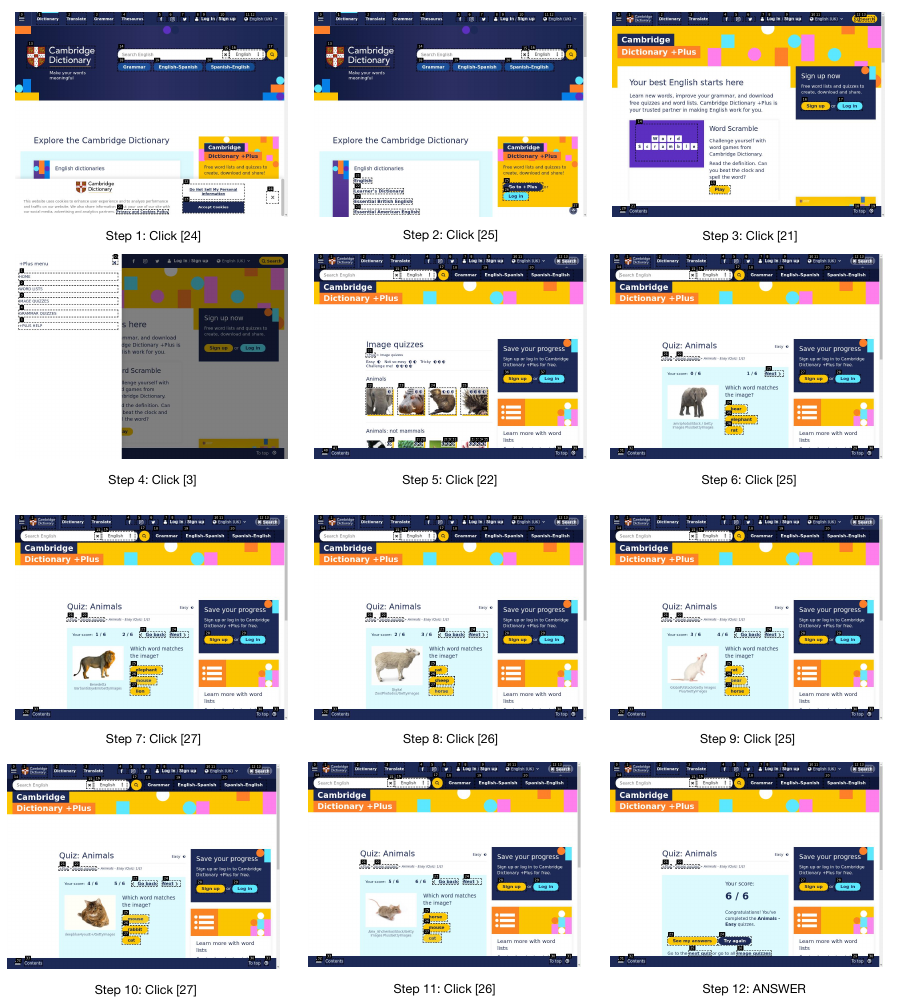}
\caption{Screenshots of a complete trajectory of browsing Cambridge Dictionary. Given the task: ``Go to the Plus section of Cambridge Dictionary, find Image quizzes and do an easy quiz about Animals and tell me your final score.'' The agent interacts with the Cambridge Dictionary website and obtains the answer: ``The final score on the easy Animals quiz is 6 out of 6.''}
\label{fig:episode_cambridge}
\end{figure*}

\begin{figure*}[t!]
\centering
\includegraphics[width=1.0\linewidth]{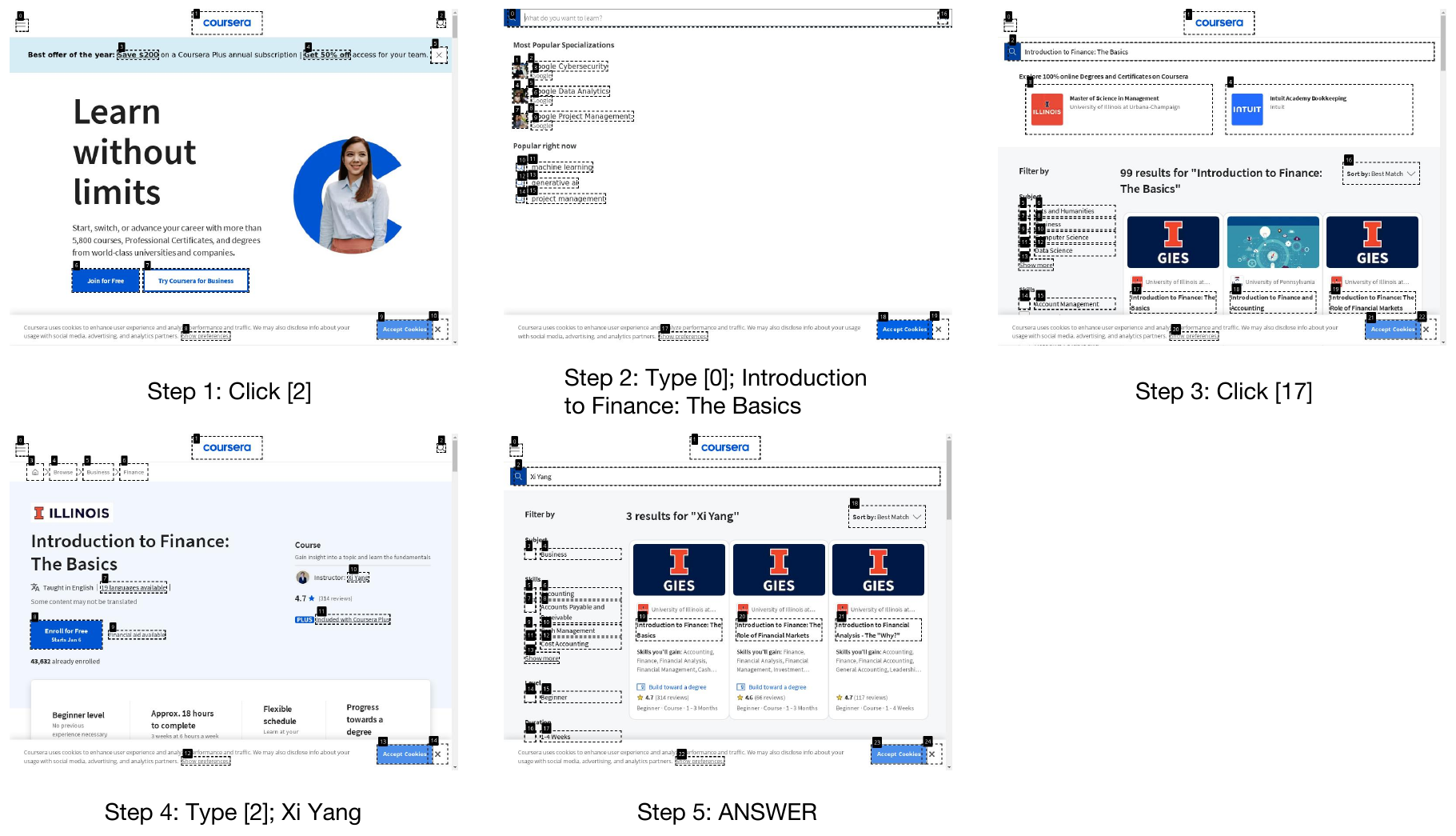}
\caption{Screenshots of a complete trajectory of browsing Coursera. Given the task: ``Identify a course on Coursera named 'Introduction to Finance: The Basics', who is the course instructor, and what other courses does he/she teach.'' The agent interacts with the Coursera website and obtains the answer: The course instructor for `Introduction to Finance: The Basics' is Xi Yang. Xi Yang also teaches `Introduction to Finance: The Role of Financial Markets' and `Introduction to Financial Analysis - The "Why?"'}
\label{fig:episode_coursera}
\end{figure*}

\begin{figure*}[t!]
\centering
\includegraphics[width=1.0\linewidth]{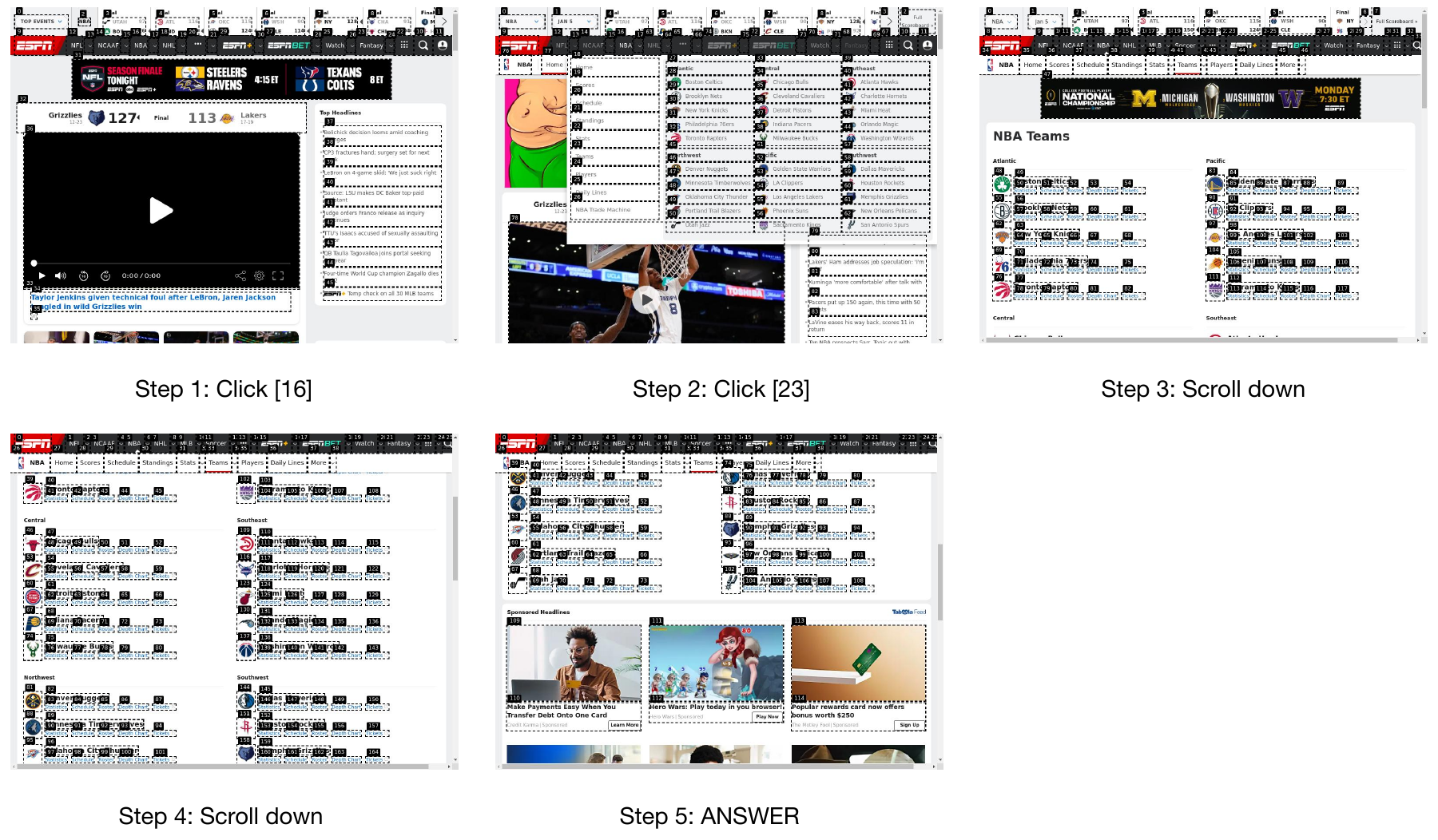}
\caption{Screenshots of a complete trajectory of browsing ESPN. Given the task: ``How many NBA teams are there and list all the teams with `New' in their name.'' The agent interacts with the ESPN website and obtains the answer: ``There are 30 NBA teams. The teams with `New' in their name are the New York Knicks and the New Orleans Pelicans.''}
\label{fig:episode_espn}
\end{figure*}

\begin{figure*}[t!]
\centering
\includegraphics[width=1.0\linewidth]{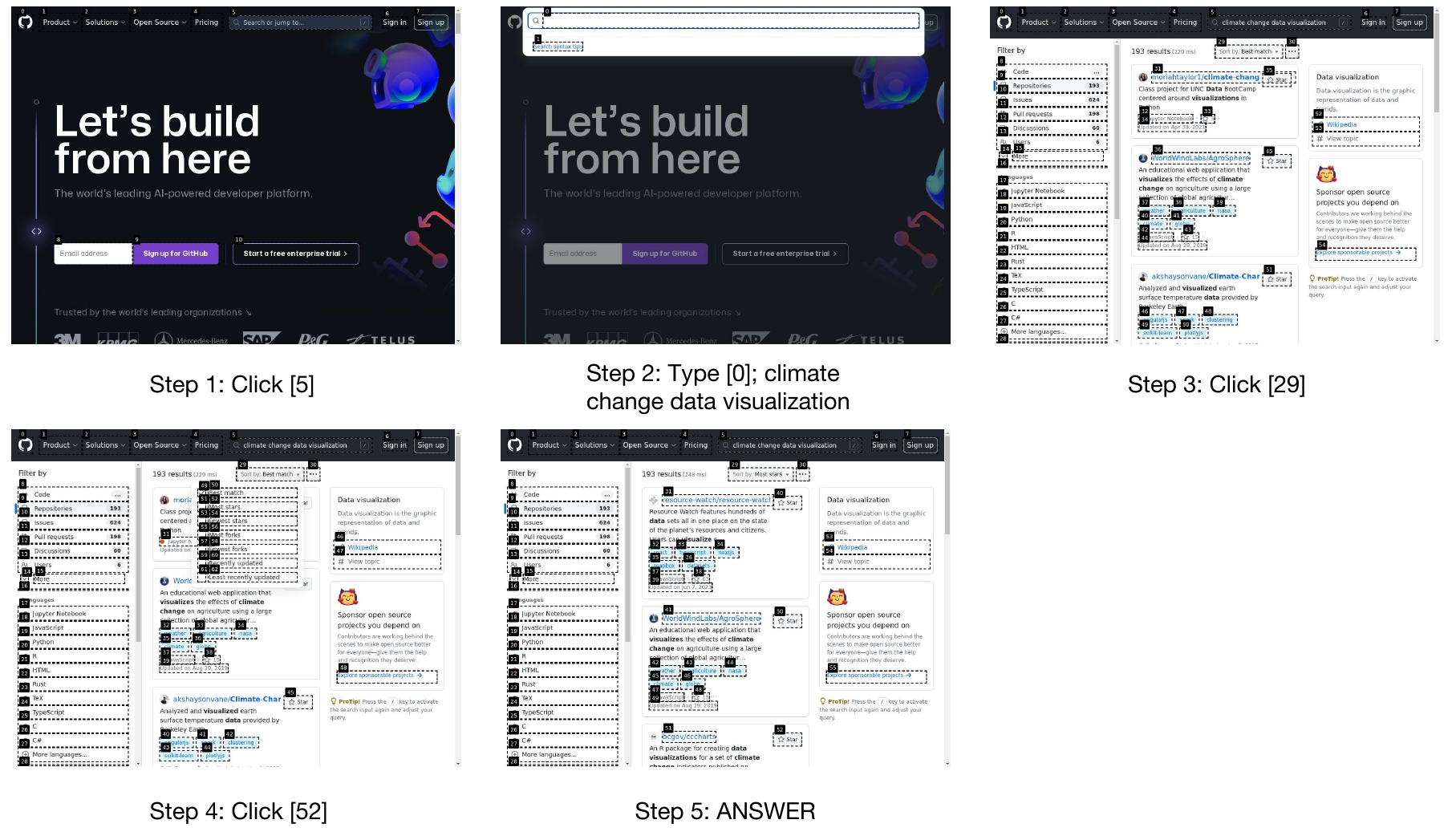}
\caption{Screenshots of a complete trajectory of browsing GitHub. Given the task: ``Search for an open-source project related to `climate change data visualization' on GitHub and report the project with the most stars.'' The agent interacts with the GitHub website and obtains the answer: ``resource-watch/resource-watch with 63 stars''}
\label{fig:episode_github}
\end{figure*}

\begin{figure*}[t!]
\centering
\includegraphics[width=1.0\linewidth]{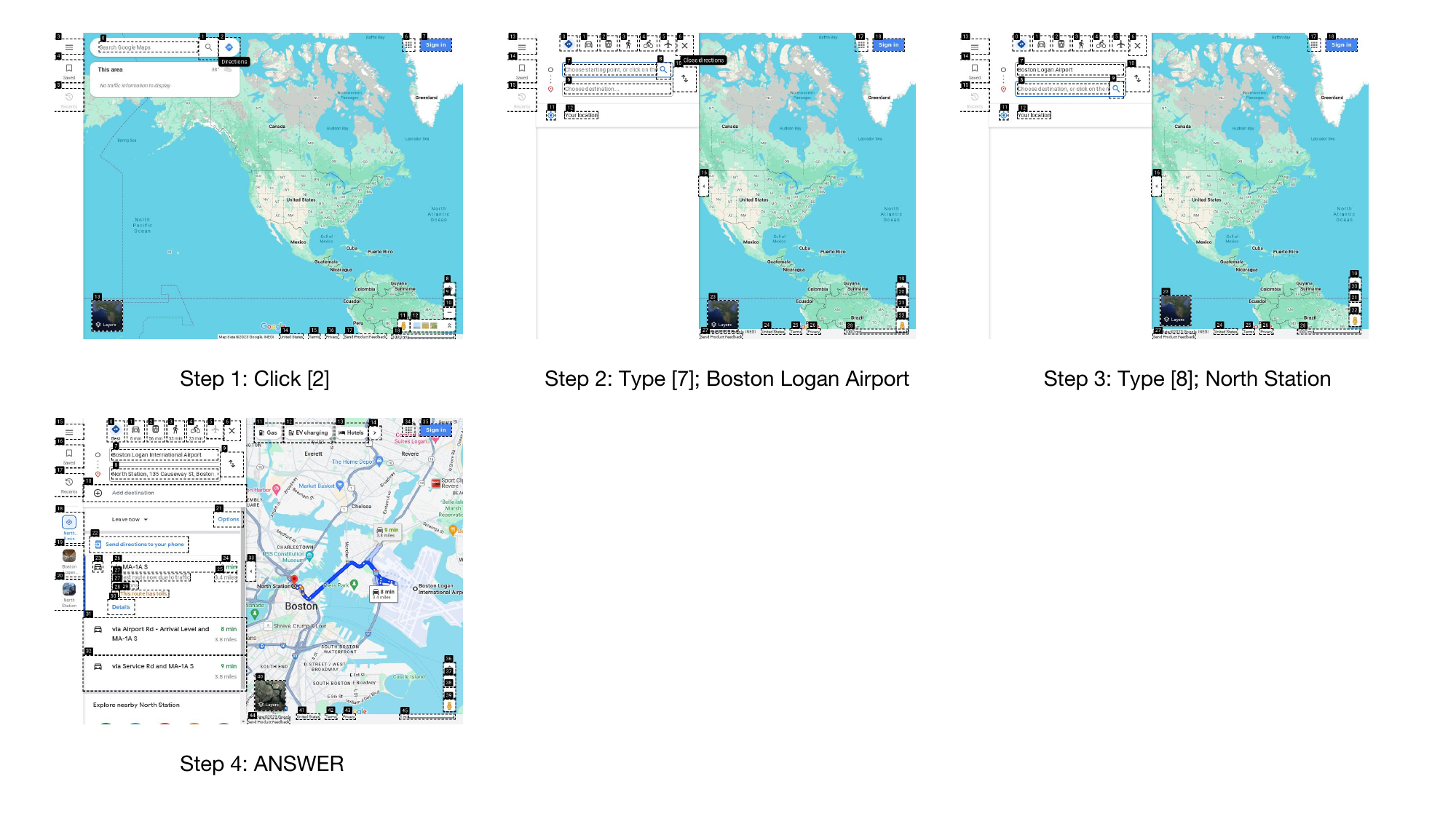}
\caption{Screenshots of a complete trajectory of browsing Google Map. Given the task: ``Plan a trip from Boston Logan Airport to North Station.'' The agent interacts with the Google Map website and obtains the answer: ``The fastest route from Boston Logan International Airport to North Station is via MA-1A S and takes approximately 8 minutes in current traffic conditions.''}
\label{fig:episode_map}
\end{figure*}

\begin{figure*}[t!]
\centering
\includegraphics[width=1.0\linewidth]{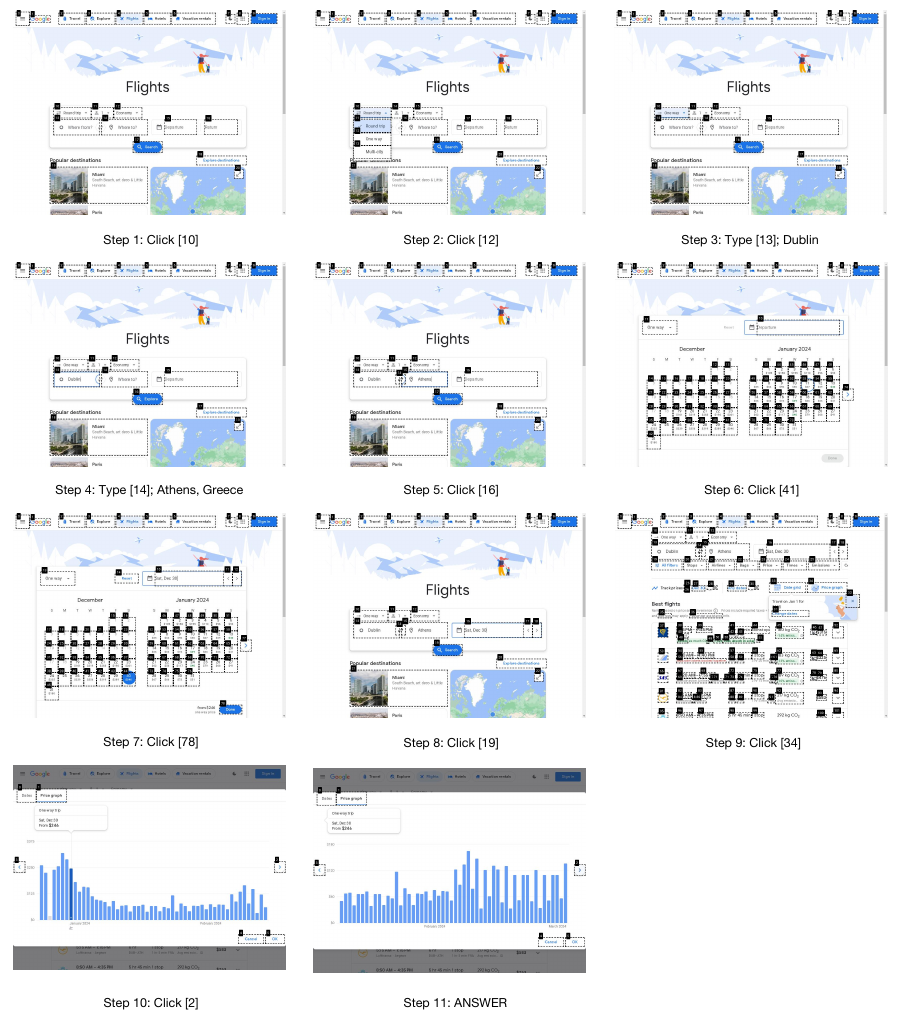}
\caption{Screenshots of a complete trajectory of browsing Google Flights. Given the task: ``Search a one-way flight from Dublin To Athens Greece for 1 Adult that leaves on December 30 and analyse the price graph for the next 2 months.'' The agent interacts with the Google Flights website and obtains the answer: ``The price graph has been analyzed, and it shows the trends for flight prices from Dublin to Athens over the next two months.''}
\label{fig:episode_flights}
\end{figure*}

\begin{figure*}[t!]
\centering
\includegraphics[width=1.0\linewidth]{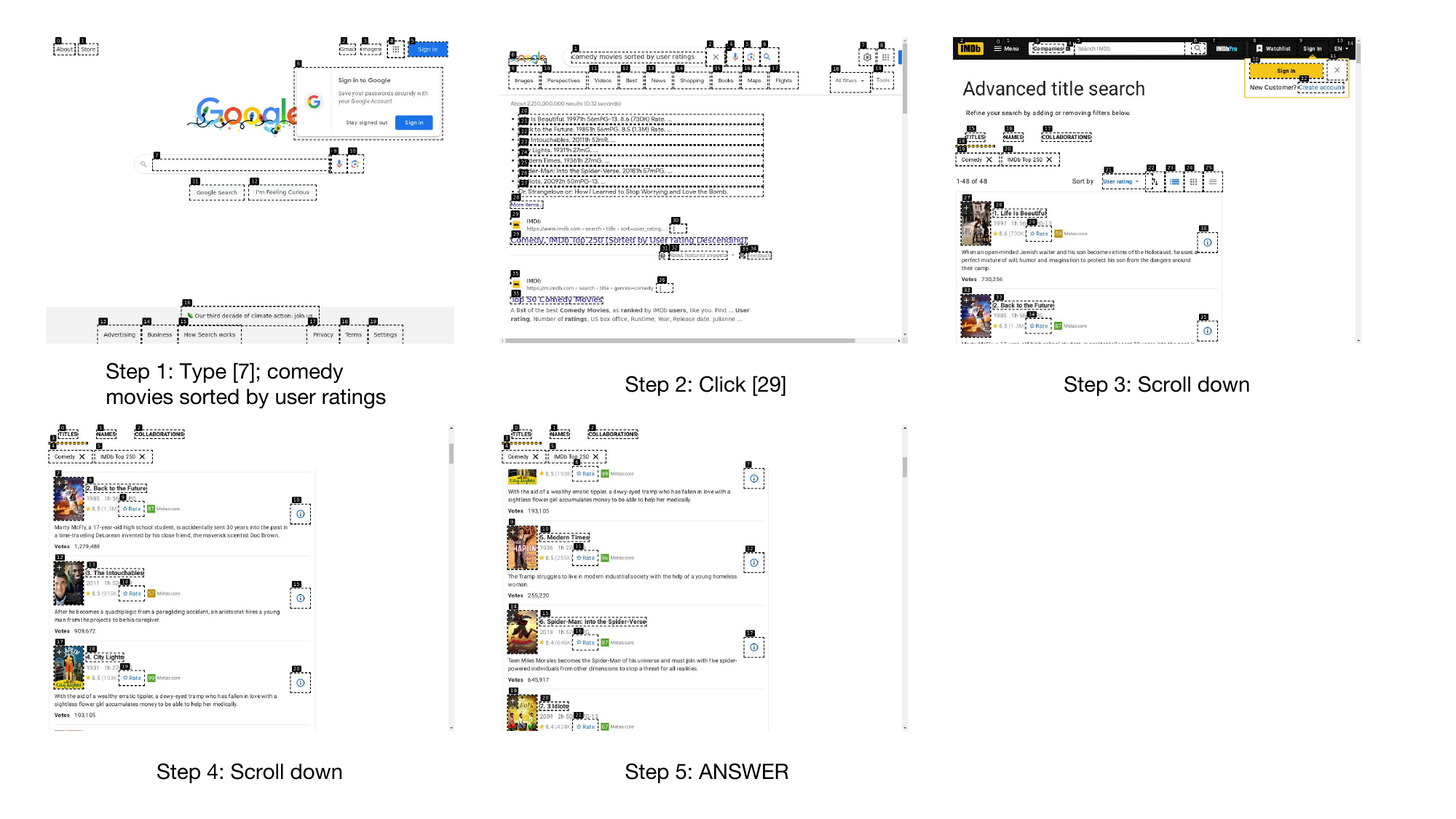}
\caption{Screenshots of a complete trajectory of browsing Google Search. Given the task: ``Show me a list of comedy movies, sorted by user ratings. Show me the Top 5 movies.'' The agent interacts with the Google Search website and obtains the answer: ``The Top 5 comedy movies sorted by user ratings are:Life Is Beautiful; Back to the Future; The Intouchables; City Lights; Modern Times''}
\label{fig:episode_google_search}
\end{figure*}

\begin{figure*}[t!]
\centering
\includegraphics[width=1.0\linewidth]{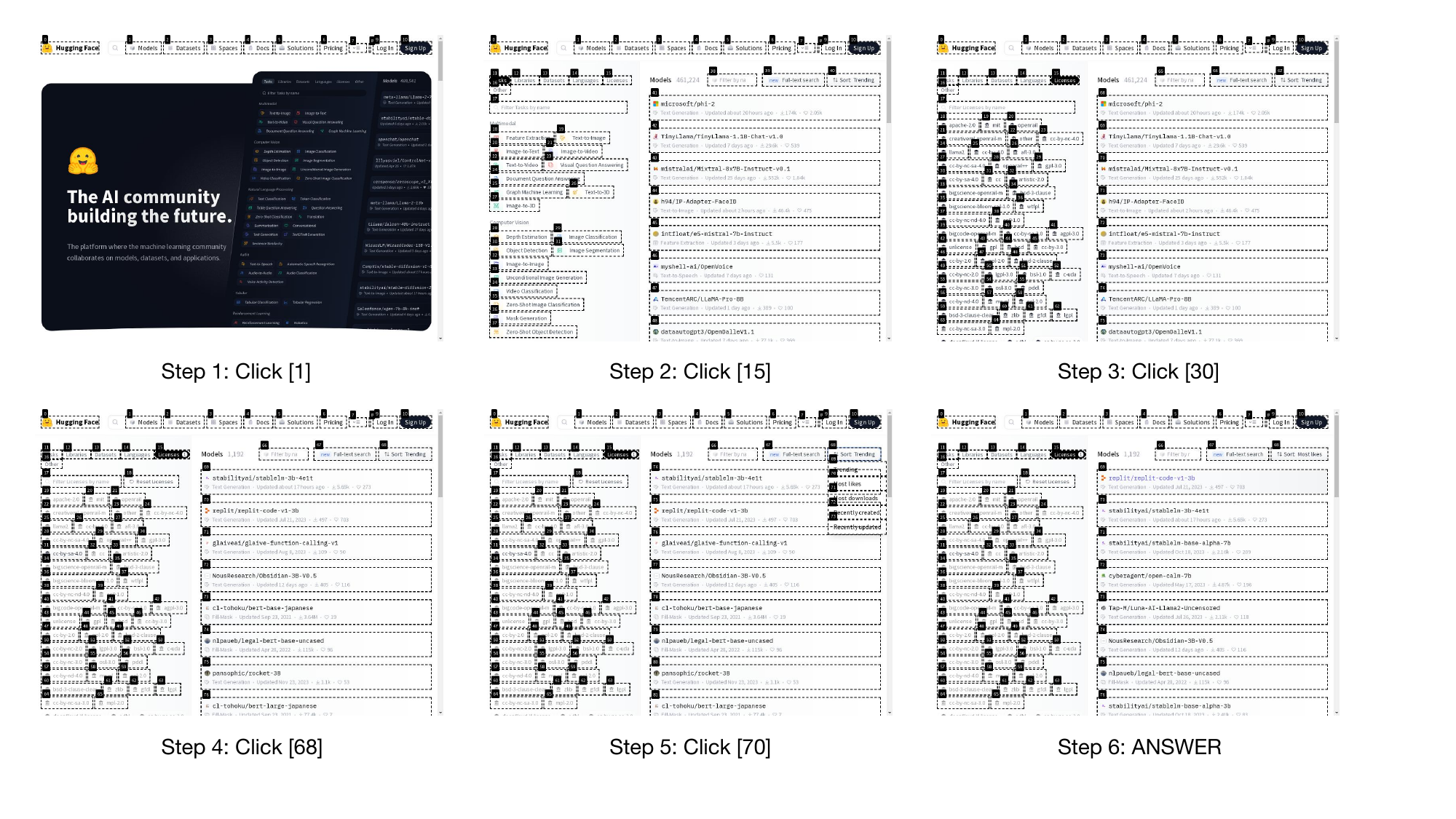}
\caption{Screenshots of a complete trajectory of browsing Huggingface. Given the task: ``Look up a model with a license of cc-by-sa-4.0 with the most likes on Huggingface.'' The agent interacts with the Huggingface website and obtains the answer: ``The model with a license of cc-by-sa-4.0 with the most likes on Hugging Face is "replit/replit-code-v1-3b" with 703 likes.''}
\label{fig:episode_huggingface}
\end{figure*}

\begin{figure*}[t!]
\centering
\includegraphics[width=1.0\linewidth]{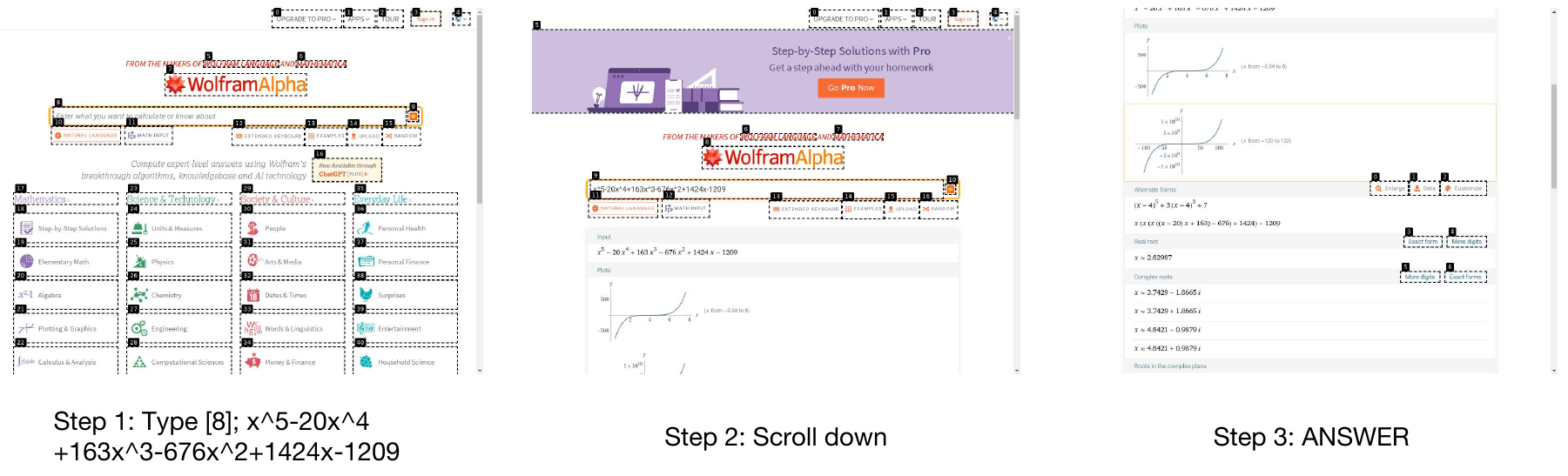}
\caption{Screenshots of a complete trajectory of browsing Wolfram Alpha. Given the task: ``Simplify x\textasciicircum5-20x\textasciicircum4+163x\textasciicircum3-676x\textasciicircum2+1424x-1209 so that it has fewer items.'' The agent interacts with the Wolfram Alpha website and obtains the answer: ``(x - 4)\textasciicircum5 + 3(x - 4)\textasciicircum3 + 7''}
\label{fig:episode_wolfram}
\end{figure*}

\begin{figure*}[t!]
\centering
\includegraphics[width=1.0\linewidth]{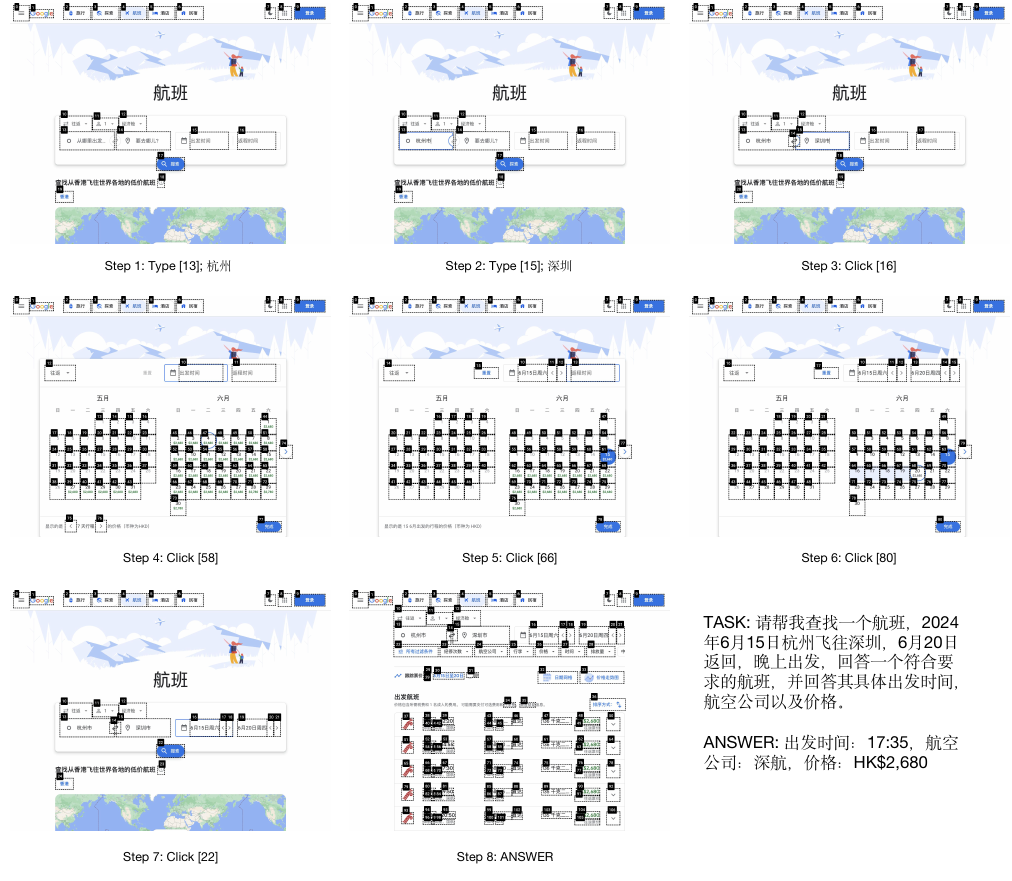}
\caption{Screenshots of a complete trajectory of browsing Google Flights in Chinese. Given the task: ``Find a flight from Hangzhou to Shenzhen on June 15, 2024, returning on June 20, departing at night, answer a flight that meets the requirements, and answer its specific departure time, airline and price.'' The agent interacts with the Google Flights website and obtains the answer: ``Departure time: 17:35, airline: Shenzhen Airlines, price: HK\$2,680''}
\label{fig:episode_flights_chinese}
\end{figure*}

\begin{figure*}[t!]
\centering
\includegraphics[width=1.0\linewidth]{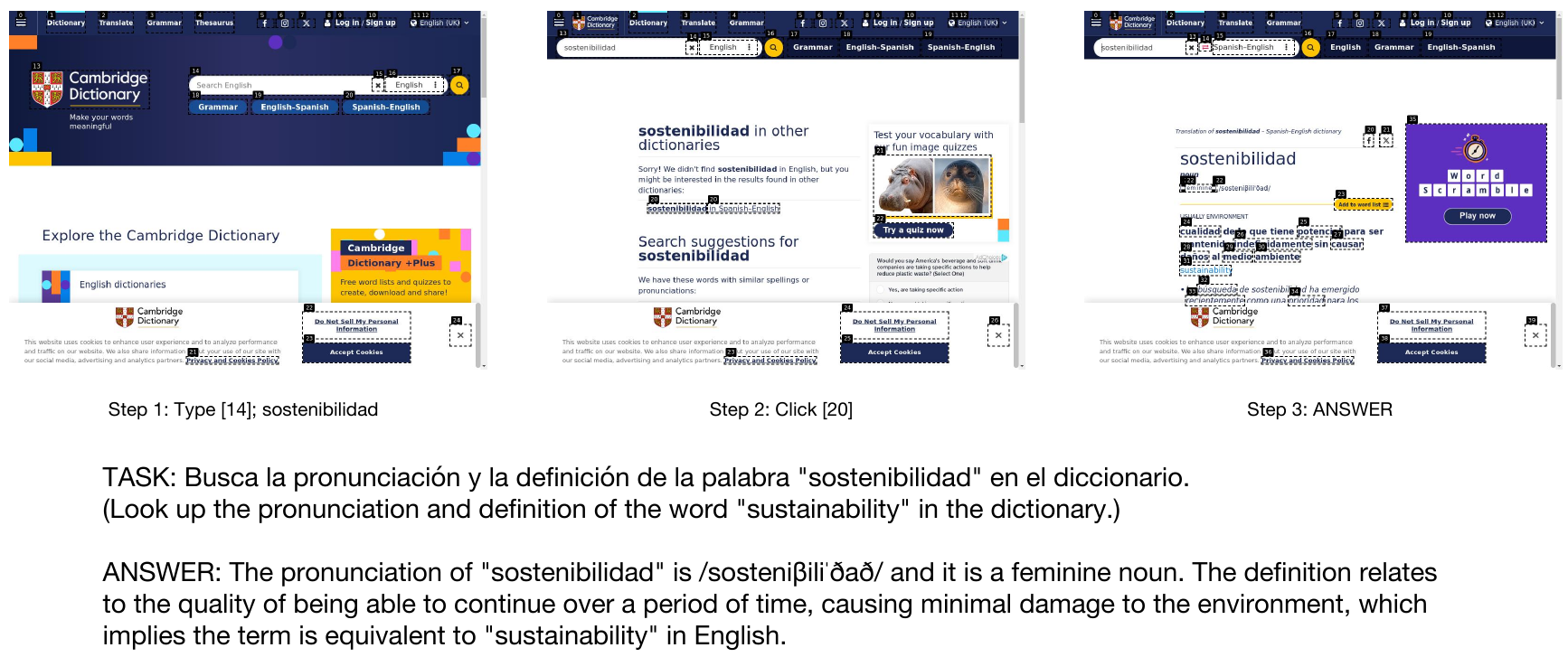}
\caption{Screenshots of a complete trajectory of browsing Cambridge Dictionary in Spanish. The description of task and answer are shown in Figure.}
\label{fig:episode_cambridge_spanish}
\end{figure*}

In Figure \ref{fig:episode}, we demonstrate how WebVoyager automatically navigates the Apple website and accomplishes the task.
In this section, we exhibit the trajectories for the remaining websites that complete web tasks successfully. We provide a screenshot for each step, accompanied by the action generated by WebVoyager. The specific navigation trajectories for each website are illustrated in Figures \ref{fig:episode_allrecipes} to \ref{fig:episode_wolfram}. In addition, we also explore the performance of WebVoyager on websites in other languages, and we provide two example trajectories in Chinese and Spanish in Figure \ref{fig:episode_flights_chinese} and Figure \ref{fig:episode_cambridge_spanish}.

\section{Additional Related Work}
\textbf{Vision-based Agents} Concurrent to our work, a few related works also studied vision-based autonomous agents. VisualWebArena \cite{koh2024visualwebarena} extends WebArena with additional websites and tasks that focus on visual reasoning to facilitate research on vision-based web agents. SeeClick \cite{cheng2024seeclick} focused on finetuning an LMM to solely leverage screenshots as inputs to interact with websites. WebVLN \cite{chen2023webvln} introduced a web simulator that provides both HTML text and screenshots to finetune supervised vision-language models. GPT-4V Wonderland \cite{yan2023gpt4v} and AppAgent \cite{zhang2023appagent} instead focus on building agents that can operate smartphone apps using the GPT-4V as the backbone. These works further underscore the promising prospects in this field.
 
\noindent \textbf{Large Multimodal Models}.
In recent years, significant strides have been made in unifying image and text representations within a single multimodal model through joint training with image and text \citep{ li2019visualbert, dosovitskiy2020image, wang2021simvlm, dai2022one, aghajanyan2022cm3}. 
Large Multimodal Models (LMMs), following in the footsteps of Large Language Models \citep{brown2020language, chen2021evaluating, chowdhery2023palm}, attain the capability of instruction following \citep{ouyang2022training} and exhibit robust multimodal comprehension. Represented by GPT-4V \citep{openai2023gpt4} and Gemini \citep{team2023gemini}, LMMs have demonstrated impressive performance on benchmarks \citep{goyal2017making, lu2022learn, zellers2019recognition, hessel2022abduction}, establishing a foundation for the construction of multimodal agents in subsequent research. 

\section{Error Cases}\label{sec:error_eg}
\begin{figure*}[t!]
\centering
\includegraphics[width=1.0\linewidth]{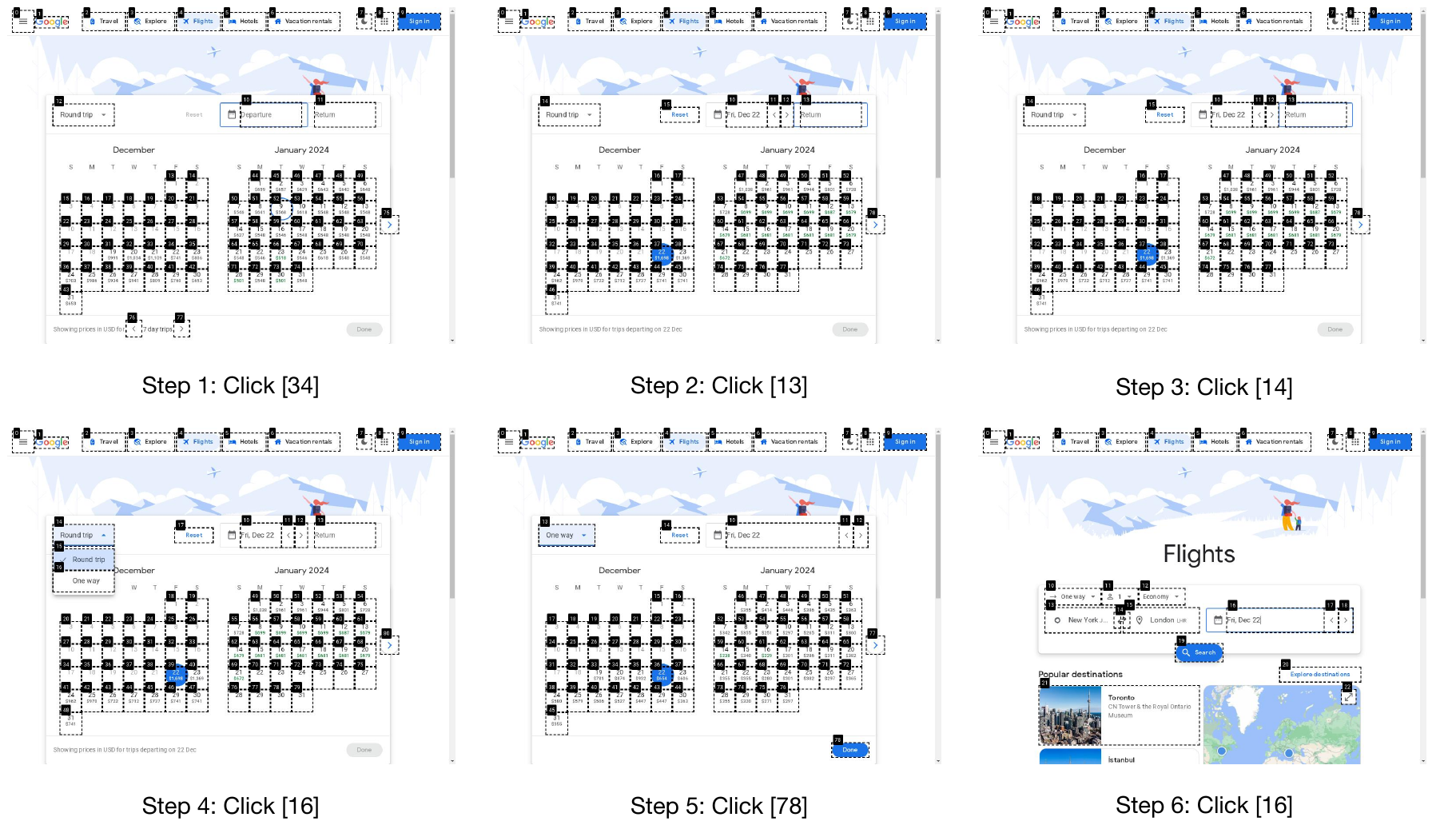}
\caption{An error case for Google Flights. Given the task:``Find the lowest fare from all eligible one-way flights for 1 adult from JFK to Heathrow on Jan. 22.'' Agent fails to select the correct numerical label though it really wants to select 22 January. }
\label{fig:error_flights}
\end{figure*}

\begin{figure*}[t!]
\centering
\includegraphics[width=1.0\linewidth]{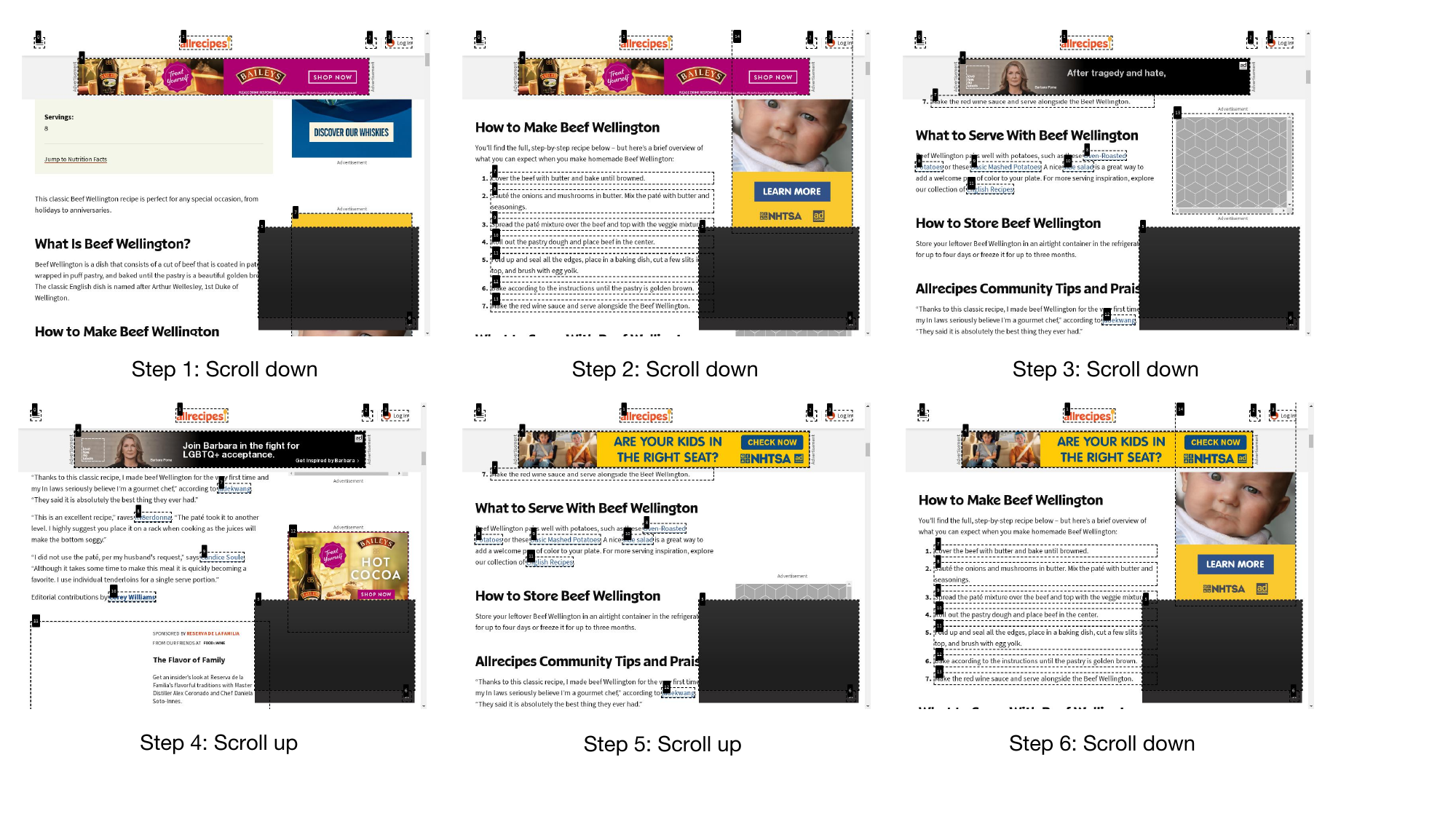}
\caption{An error case for Allrecipes. Given the task:``Search for a recipe for Beef Wellington on Allrecipes that has at least 200 reviews and an average rating of 4.5 stars or higher. List the main ingredients required for the dish.'' Agent fails to scroll the page correctly and find ingredients.}
\label{fig:error_allrecipes}
\end{figure*}

\begin{figure*}[t!]
\centering
\includegraphics[width=1.0\linewidth]{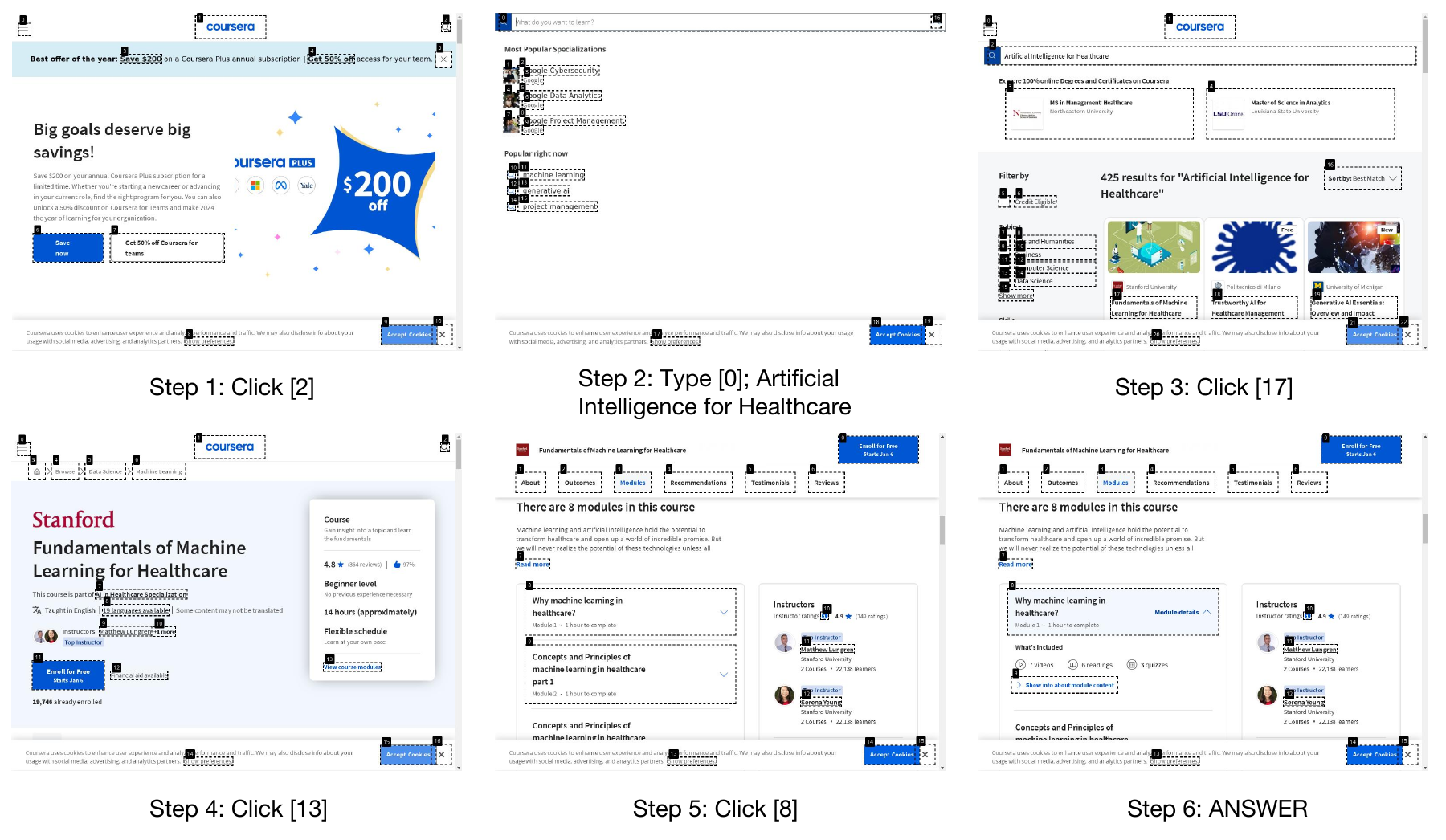}
\caption{An error case for Coursera. Given the task: ``Identify a course on Coursera related to 'Artificial Intelligence for Healthcare' and note the course duration along with the number of quizzes in Assessments.'' The agent provides the answer: ``Module 1 includes 3 quizzes in the Assessments.'' However, we also require other Modules and agent fails to provide the results.}
\label{fig:error_coursera}
\end{figure*}

\begin{figure*}[t!]
\centering
\includegraphics[width=1.0\linewidth]{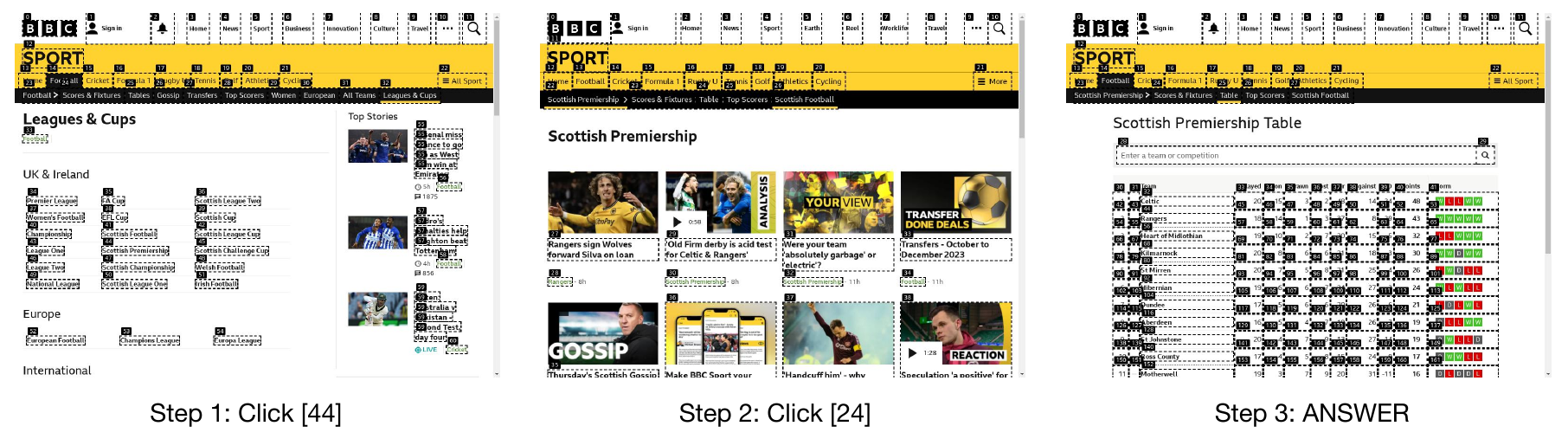}
\caption{An error case for BBC News. Given the task: ``Find out how many teams are in the Scottish Premiership of the Football Tournament and when did the Hibernian team's most recent match start?'' The agent provides the answer: ``There are 12 teams in the Scottish Premiership. To find out the exact start time of Hibernian's most recent match, further interaction with the website would be required.'' The Agent knows that the task is not yet complete, but it ends its navigation early, even though it can find the Hibernian team's most recent match.}
\label{fig:error_bbc}
\end{figure*}

In this section, we provide specific examples of the four types of errors mentioned in the Error Analysis section. Figure \ref{fig:error_flights} illustrates a snippet of WebVoyager navigating on Google Flights and a Visual Grounding Issue appears. The task is to retrieve one-way flight information for January 22nd; however, it selects December 22nd on the Calendar and fails to make the necessary corrections. Although it attempts to modify the date in step 6, it ultimately fails to do so. Figure \ref{fig:error_allrecipes} illustrates a situation of WebVoyager navigating on Allrecipes, encountering the Navigation Stuck issue. The agent requires multiple downward scrolls to locate the correct ingredients. However, it experiences confusion during the process, and it is uncertain whether to scroll up or down. Figure \ref{fig:error_coursera} depicts the Hallucination issue encountered by WebVoyager on the Coursera website. In the task, we query the number of quizzes in the "Artificial Intelligence for Healthcare" course. However, the agent only identifies the quiz in module 1 of the course, which is not the optimal answer and does not fulfill the task requirements. Figure \ref{fig:error_bbc} illustrates the issue of Prompt Misalignment encountered while browsing BBC News. WebVoyager fails to fulfill all the task requirements. Instead of completing the navigation, it provides partial answers and tells me how to find complete answers, which is not end-to-end.

\end{document}